\pdfoutput=1
\documentclass[11pt]{article}

\usepackage[group-separator={,}]{siunitx}
\usepackage{booktabs} % For formal tables
\usepackage{color}
\usepackage{listings}
\usepackage{enumitem}
\usepackage{threeparttable}
\usepackage{multirow}% http://ctan.org/pkg/multirow
\usepackage{hhline}% http://ctan.org/pkg/hhline
\usepackage{subfigure}
\usepackage{array}
\usepackage{pifont}
\usepackage{booktabs}
\usepackage{dsfont}
\usepackage{microtype}
\usepackage{graphicx}
\usepackage{stmaryrd}
\usepackage{amsmath}
\usepackage{amsthm}
\usepackage{amssymb}
\usepackage{url}
\usepackage{bbm}
\usepackage{xspace}
\usepackage[normalem]{ulem}
\usepackage{booktabs, siunitx}
\usepackage[svgnames,table]{xcolor}
\usepackage[tableposition=above]{caption}
\usepackage{xcolor}
\usepackage{tkz-euclide}
\usepackage[framemethod=tikz]{mdframed}
\usepackage{empheq}
\usepackage[most]{tcolorbox}
\usepackage{adjustbox}
\usepackage{wrapfig}
\usepackage{hyperref}
\usepackage[capitalize]{cleveref}

\usepackage{balance}
\usepackage[linesnumbered,ruled,vlined]{algorithm2e}
\expandafter\def\expandafter\UrlBreaks\expandafter{\UrlBreaks%  save the current one
  \do\a\do\b\do\c\do\d\do\e\do\f\do\g\do\h\do\i\do\j%
  \do\k\do\l\do\m\do\n\do\o\do\p\do\q\do\r\do\s\do\t%
  \do\u\do\v\do\w\do\x\do\y\do\z\do\A\do\B\do\C\do\D%
  \do\E\do\F\do\G\do\H\do\I\do\J\do\K\do\L\do\M\do\N%
  \do\O\do\P\do\Q\do\R\do\S\do\T\do\U\do\V\do\W\do\X%
  \do\Y\do\Z}

\usepackage{amsmath,amsfonts,bm}

\def\Figref#1{Fig~\ref{#1}}
\def\Twofigref#1#2{Figures \ref{#1} and \ref{#2}}

\def\Secref#1{Section~\ref{#1}}
\def\Tblref#1{Table~\ref{#1}}
\def\eqref#1{eq~\ref{#1}}
\def\Eqref#1{Eq~\ref{#1}}
\def\Lemmaref#1{Lemma~\ref{#1}}
\def\Theref#1{Theorem~\ref{#1}}
\def\1{\bm{1}}

\DeclareMathAlphabet{\mathsfit}{\encodingdefault}{\sfdefault}{m}{sl}
\SetMathAlphabet{\mathsfit}{bold}{\encodingdefault}{\sfdefault}{bx}{n}

\def\gX{{\mathcal{X}}}
\def\gY{{\mathcal{Y}}}

\newcommand{\E}{\mathbb{E}}

\newcommand{\R}{\mathbb{R}}

\DeclareMathOperator*{\argmin}{arg\,min}

\newcommand{\cmark}{\ding{51}}%
\newcommand{\xmark}{\ding{55}}%

\expandafter\def\expandafter\UrlBreaks\expandafter{\UrlBreaks%  save the current one
  \do\a\do\b\do\c\do\d\do\e\do\f\do\g\do\h\do\i\do\j%
  \do\k\do\l\do\m\do\n\do\o\do\p\do\q\do\r\do\s\do\t%
  \do\u\do\v\do\w\do\x\do\y\do\z\do\A\do\B\do\C\do\D%
  \do\E\do\F\do\G\do\H\do\I\do\J\do\K\do\L\do\M\do\N%
  \do\O\do\P\do\Q\do\R\do\S\do\T\do\U\do\V\do\W\do\X%
  \do\Y\do\Z}

\definecolor{codegreen}{rgb}{0,0.6,0}
\definecolor{codegray}{rgb}{0.5,0.5,0.5}
\definecolor{codepurple}{rgb}{0.58,0,0.82}
\definecolor{backcolour}{rgb}{1,1,1}
 
\lstdefinestyle{mystyle}{
    backgroundcolor=\color{backcolour},   
    commentstyle=\color{codegreen},
    keywordstyle=\color{magenta},
    numberstyle=\tiny\color{codegray},
    stringstyle=\color{codepurple},
    basicstyle=\ttfamily\footnotesize,
    breakatwhitespace=false,         
    breaklines=true,                 
    captionpos=b,                    
    keepspaces=true,                 
    numbers=left,                    
    numbersep=5pt,                  
    showspaces=false,                
    showstringspaces=false,
    showtabs=false,           
    tabsize=2,
    xleftmargin=1.5em
}

\newtheorem{lemma}{Lemma}[section]
\newtheorem{theorem}{Theorem}[section]

\newtheorem{example}{Example}[section]

\newcommand{\hstate}[0]{h}

\newcommand{\spec}[0]{S}

\newcommand{\alphabet}[0]{\Sigma}
\newcommand{\bfx}[0]{\mathbf{x}}
\newcommand{\bfz}[0]{\mathbf{z}}
\newcommand{\bft}[0]{\mathbf{t}}

\def\substring#1#2#3{#1_{#2:#3}}

\def\lb#1{{#1}^\emph{(lb)}}

\newcommand{\toolname}[0]{ARC}
\newcommand{\reviewperturb}[0]{S_\emph{review}}
\newcommand{\jiaapproach}[0]{CertSub}
\newcommand{\Tinsword}{T_{\emph{Dup}}}
\newcommand{\Tdelword}{T_{\emph{DelStop}}}
\newcommand{\Tsubword}{T_{\emph{SubSyn}}}
\newcommand{\SDS}{\{(\Tdelword{}, 2), (\Tsubword{}, 2)\}}%{S_\emph{DS}}
\newcommand{\SIS}{\{(\Tinsword{}, 2), (\Tsubword{}, 2)\}}%{S_\emph{IS}}
\newcommand{\SDI}{\{(\Tdelword{}, 2), (\Tinsword{}, 2)\}}%{S_\emph{DI}}

\newcommand{\lengththreecell}{4.3cm}
\newcommand{\lengthcell}{0.82cm}
\newcommand{\lengthcellshorter}{0.6cm}
\newcommand{\lengthcelllonger}{0.84cm}
\newcommand{\Exhaustiveshort}{EX Acc.}
\newcommand{\Verifiedshort}{CF Acc.}

\newcommand{\lstm}{\textsc{lstm}}
\newcommand{\treelstm}{\textsc{trlstm}}
\newcommand{\ibp}[1]{\widehat{#1}}

\newcommand{\tightperturb}{S^=}
\newcommand{\tightperturbset}[1]{\textsc{dec}_{#1}}

\renewcommand{\leq}{\leqslant}
\renewcommand{\geq}{\geqslant}
\newcommand{\perturbed}[1]{{\color{red}{#1}}}
\newcommand{\len}[1]{\textsc{len}_{#1}}
\newcommand{\maxlen}[1]{\textsc{maxlen}_{#1}}
\newcommand{\size}[1]{\textsc{size}_{#1}}
\newcommand{\reverse}{\emph{R}}

\newcommand{\TReviewFirst}[0]{T_{\emph{review}1}}
\newcommand{\TReviewSecond}[0]{T_{\emph{review}2}}
\newcommand{\TReviewThird}[0]{T_{\emph{review}3}}
\newcommand{\TReviewFourth}[0]{T_{\emph{review}4}}

\usepackage{emnlp2021}
\usepackage{arabtex}
\renewenvironment{abstract}%
         {\centerline{\large\bf Abstract}%
          \begin{list}{}%
             {\setlength{\rightmargin}{0.6cm}%
              \setlength{\leftmargin}{0.6cm}}%
           \item[]\ignorespaces}%
         {\unskip\end{list}}
         
\usepackage{times}
\usepackage{latexsym}

\usepackage[T1]{fontenc}
\usepackage[utf8]{inputenc}

\usepackage{microtype}

\renewcommand{\paragraph}[1]{\vspace{.5em}\noindent\textbf{#1.}}

\title{Certified Robustness to Programmable Transformations in LSTMs}

\author{Yuhao Zhang, Aws Albarghouthi\thanks{~Author's name in native alphabet: \novocalize\RL{'aws albr.gU_ty}}, Loris D'Antoni \\
        Department of Computer Science, University of Wisconsin-Madison, USA. \\ \{yuhaoz, aws, loris\}@cs.wisc.edu}
\newtcbox{\mymath}[1][]{%
    nobeforeafter, math upper, tcbox raise base,
    enhanced, colframe=black,
    colback=white!30, boxrule=1pt,
    #1}

\begin{document}
\maketitle

\begin{abstract}
Deep neural networks for  natural language processing
are fragile in the face of adversarial examples---%
small input perturbations, like synonym substitution or word duplication, which  cause a neural network to change its prediction.
%
% We present an approach to certifying the robustness of, and training robust, LSTMs (and extensions of LSTMs). 
We present an approach to \emph{certifying} the robustness of LSTMs (and extensions of LSTMs) and \emph{training} models that can be efficiently certified. 
Our approach can certify robustness to intractably large perturbation spaces defined \emph{programmatically} in a language of string transformations.
%
% The key insight of our approach is an application of \emph{abstract interpretation} 
% that exploits  recursive LSTM structure  to incrementally propagate symbolic sets of inputs,  compactly representing a large perturbation space.
% We train models using our approach on standard datasets and show that the resulting models have high certification accuracy to combinations of meaning-preserving transformations. 
Our evaluation shows that (1) our approach can train models that are more robust to combinations of string transformations than those produced using existing techniques; (2) our approach can show high certification accuracy of the resulting models.
\end{abstract}

% \aws{I've modified the paragraph definition; it automatically adds a period; make sure there are no doulbe periods}
\section{Introduction}
Adversarial examples are small perturbations of an input that fool a neural network into changing its prediction~\cite{DBLP:conf/ccs/Carlini017, DBLP:journals/corr/SzegedyZSBEGF13}.
In NLP, 
adversarial examples involve modifying an input string
by, for example, replacing words with synonyms, deleting stop words, inserting words, etc.~\cite{ebrahimi2018hotflip, DBLP:conf/ndss/LiJDLW19, DBLP:conf/acl/ZhangZML19}.
% Adversarial examples rise security concerns for all real-world applications that run on DNN.

% Over the past few years, researchers have developed a promising arsenal of \emph{defense} techniques against adversarial examples in NLP tasks, resulting in models that are increasingly \emph{robust} to adversarial inputs. 
Ideally, a \emph{defense}  against adversarial examples in NLP tasks
should fulfill the following desiderata: (1) Handle \textbf{recursive models}, like LSTMs and extensions thereof, which are prevalent in NLP. (2) Construct \textbf{certificates} (proofs) of robustness. (3) Defend against \textbf{arbitrary string transformations}, like combinations of word deletion, insertion, etc.

\begin{figure}[t]
    \centering
    \def\drawcubeat#1#2#3#4#5#6#7{
\pgfmathsetmacro{\cubex}{#1}
\pgfmathsetmacro{\cubey}{#2}
\pgfmathsetmacro{\cubez}{#3}
\pgfmathsetmacro{\startx}{#4}
\pgfmathsetmacro{\starty}{#5}
\pgfmathsetmacro{\startz}{#6}
\draw[black,fill=yellow,opacity=0.4] (\startx,\starty,\startz) -- ++(-\cubex,0,0) -- ++(0,-\cubey,0) -- ++(\cubex,0,0) -- cycle;
\draw[black,fill=yellow,opacity=0.4] (\startx,\starty,\startz) -- ++(0,0,-\cubez) -- ++(0,-\cubey,0) -- ++(0,0,\cubez) -- cycle;
\draw[black,fill=yellow,opacity=0.4] (\startx,\starty,\startz) -- ++(-\cubex,0,0) -- ++(0,0,-\cubez) -- ++(\cubex,0,0) -- cycle;
\node at (\startx - \cubex * 0.5 + \cubez * 0.15,\starty + \cubez * 0.15 + 0.05) {#7};
}

\def\drawlstm#1#2#3#4#5#6{ % LSTMtext, id of LSTM node, (x, y), word, id of text
\pgfmathsetmacro{\x}{#3}
\pgfmathsetmacro{\y}{#4}
\node[draw, thick, rounded corners=5] (#2) at (\x,\y) {#1};
\node[draw=red!0,fill=yellow!40] (#6) at (\x,\y + 0.6) {#5};
\draw[->] (#6) -- (#2);
}

\def\drawsinglelstm#1#2#3#4{ % LSTMtext, id of LSTM node, (x, y)
\pgfmathsetmacro{\x}{#3}
\pgfmathsetmacro{\y}{#4}
\node[draw, thick, rounded corners=5] (#2) at (\x,\y) {#1};
\draw[<-] (#2) -- ++(0, 0.4);
}

\trimbox{-0.7cm 0.5cm 0cm 0cm}{ 
\begin{tikzpicture}[scale=1]
	\tikzstyle{every node}=[font=\small]
% 	\draw[help lines,step=5mm,gray!20] (-0.5,0) grid (7,-4);

    %%%%%%%%%% lstms %%%%%%%%%% 
    \pgfmathsetmacro{\xa}{0}
    \pgfmathsetmacro{\xb}{1.9}
    \pgfmathsetmacro{\xc}{3.8}
    \pgfmathsetmacro{\xd}{5.2}
    \pgfmathsetmacro{\xmid}{2.2}
    \pgfmathsetmacro{\yc}{-1.1}
    \pgfmathsetmacro{\yd}{-2.9}
    \pgfmathsetmacro{\ye}{-3.4}
    
	\drawlstm{$\lstm$}{A1}{\xa}{-2}{to}{A2};
	\drawlstm{$\lstm$}{B1}{\xb}{-2}{the}{B2};
	\drawlstm{$\lstm$}{C1}{\xb}{\ye}{movie}{C2};
	\drawlstm{$\lstm$}{D1}{\xc}{\yc}{movie}{D2};
	\drawlstm{$\lstm$}{B0}{\xa}{\ye}{the}{B0_1};
	\drawsinglelstm{$\lstm$}{E1}{\xc}{\yd};
% 	\drawlstm{$\lstm$}{G1}{\xd}{-2}{...}{G2};
	
	\node (F1) at (\xc, \ye) {...}; 
	\node (F2) at (\xd, \yc) {...}; 
	\node (F3) at (\xd, \yd) {...}; 
% 	\node (F4) at (\xe, -2) {...}; 
	
	%%%%%%%%%% cubes %%%%%%%%%%
	
% 	\drawcubeat{0.3}{0.3}{0.3}{1.1}{-1.7}{0}{};
% 	\drawcubeat{0.3}{0.3}{0.3}{0.7}{-3}{0}{};
% 	\drawcubeat{0.3}{0.3}{0.3}{2.9}{-1.1}{0}{};
% 	\drawcubeat{0.3}{0.3}{0.3}{2.9}{-2.2}{0}{};
% 	\drawcubeat{0.3}{0.3}{0.3}{5.15}{-1}{0}{};
% 	\drawcubeat{0.3}{0.3}{0.3}{5.1}{-2.3}{0}{};
	
	%%%%%%%%%% abstract cube %%%%%%%%%%
	
	\pgfmathsetmacro{\xcorner}{\xc+0.55}
	\pgfmathsetmacro{\ycorner}{\yc-0.65}
	\drawcubeat{1.3}{0.7}{0.55}{\xcorner}{\ycorner}{0}{\small{\textbf{abstract}}};
	\node at (\xcorner-1.2, \ycorner-0.35) {\Large{\{}};
    \node (film) at (\xcorner-0.65, \ycorner-0.2) {film};
    \node (movies) at (\xcorner-0.65, \ycorner-0.45) {movies};
    \node at (\xcorner-0.1, \ycorner-0.35) {\Large{\}}};
    
    %%%%%%%%%% lines %%%%%%%%%%

    \draw[<-] (\xa-0.5, -2) -- ++(-0.3, 0);
    \draw[->] (A1) -- (B1);
    \draw[->] (C1) -- (F1);
    \draw[->] (D1) -- (F2);
    \draw[->] (E1) -- (F3);
    \draw[<-] (\xa-0.5, \ye) -- ++(-0.3, 0);
    \draw[->] (B0) -- (C1);
    % \draw[->] (G1) -- (F4);
    \draw[->] (A1) .. controls (\xa+1, -2) and (\xa+0.5, \ye) .. (C1);
    \draw[->] (B1) .. controls (\xb+1.3, -2) and (\xc-1, \yc) .. (D1);
    \draw[->] (B1) .. controls (\xb+1.3, -2) and (\xc-1, \yd) .. (E1);
    % \draw[->] (D1) .. controls (\xc+1, \yc) and (\xd-1.3, -2) .. (G1);
    % \draw[->] (E1) .. controls (\xc+1, \yd) and (\xd-1.3, -2) .. (G1);
    
    %%%%%%%%%%% memoize %%%%%%%%%%
    
    % \begin{scope}[opacity=0.7,dashed]
    %     \drawfadelstm{$\lstm$}{a1}{0}{-3.5}{to}{a2};
    %     \draw[<-] (-0.5, -3.5) -- ++(-0.3, 0);
    %     \draw[->] (a1) -- (C1);
    %     % \drawcubeat{0.3}{0.3}{0.3}{1}{-3.7}{0}{};
    % \end{scope}
    
    % \draw[fill=green, opacity=0.1, dashed, rounded corners=5, thick] (-0.8,-1) rectangle (0.6,-3.9);
    % \node at (-0.1,-0.8) {\textbf{\emph{memoize}}};
    
    \draw[thick, RoyalBlue, dashed] (0.5, -1.7) .. controls (0.8, -2) .. (0.5, -2.3);
    \node[text width=85, fill=gray!20] (memoize) at (1.2,-0.7) {\textbf{memoize} hidden states};
    \draw[RoyalBlue, ->, thick, >=latex] (0.6, -1.7) -- (memoize);
    
    \draw[thick, RoyalBlue, dashed] (\xb+0.5, -1.7) .. controls (\xb+0.8, -2) .. (\xb+0.5, -2.3);
    % \draw[thick, RoyalBlue, dashed] (\xc+0.5, \yc+0.3) .. controls (\xc+0.8, \yc) .. (\xc+0.5, \yc-0.3);
    % \draw[thick, RoyalBlue, dashed] (\xc+0.5, \yd+0.3) .. controls (\xc+0.8, \yd) .. (\xc+0.5, \yd-0.3);
    
    %%%%%%%%%% join @@@@@@@@@@

    \draw[thick, Green, dashed] (\xb-0.9, \ye + 0.3) .. controls (\xb-0.6, \ye) .. (\xb-0.9, \ye - 0.3);
    \node[text width=66, fill=gray!20] (join) at (\xb,-4.1) {\textbf{join} hidden states};
    \draw[Green, ->, thick, >=latex] (\xb-0.8, \ye - 0.2) -- (join);
    
    %%%%%%%%%% others %%%%%%%%%%
    
    % \node at (\xb, \ye - 0.35) {...};
    % \node at (\xc, \ye - 0.35) {...};
    % \node at (\xd, \ye - 0.35) {...};
    
    \node at (\xmid, \ye - 1.3) {(b) \toolname{}: Abstract Recursive Certification};

    %%%%%%%%%% enumerate %%%%%%%%%%
    \begin{scope}[yshift=1.5cm, on grid, mytext/.style={anchor=west, text width=160, text height=1}, myline/.style={draw=red,thick}]
	    \pgfmathsetmacro{\span}{0.29}
    	\node[mytext](A) at (\xmid-1.8,0) {to ~~~~ the ~~~~ movie}; 
    	\node[mytext, below = \span of A](B) {to ~~~~ the ~~~~ \perturbed{film}};
    	\node[mytext, below = \span of B](C) {to ~~~~ the ~~~~ \perturbed{movies}};
    	\node[mytext, below = \span of C](D) {\perturbed{\sout{to}} ~~~~ the ~~~~ movie};
    	\node[mytext, below = \span of D](E) {to ~~~~ \perturbed{\sout{the}} ~~~~ movie};
    	
    	\foreach \n in {A,B,C,D,E}
            \node[right] at (\n) {~~~~ ...};

	    \node at (\xmid, -1.5) {(a) Enumeration of all strings};
    \end{scope}
\end{tikzpicture}
}
    \caption{An illustration of our approach.}
    \label{fig:approachfig}
\end{figure}
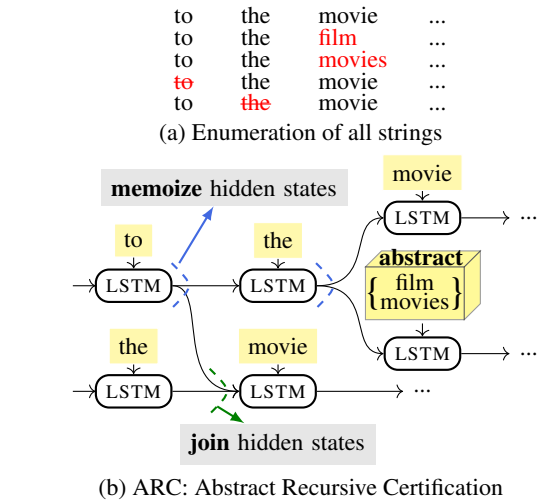

It is quite challenging to fulfill all three desiderata;
indeed, existing techniques are forced to make tradeoffs.
For instance,
the theoretical insights underlying a number of certification approaches are intimately tied to symbol substitution~\cite{DBLP:conf/emnlp/JiaRGL19,DBLP:conf/emnlp/HuangSWDYGDK19,DBLP:conf/acl/YeGL20, DBLP:conf/nips/XuS0WCHKLH20, dong2021towards},
and some techniques cannot handle recursive models~\cite{DBLP:conf/emnlp/HuangSWDYGDK19, ZhangAD20}.
On the other hand, techniques that strive to be robust
to arbitrary string transformations achieve this  at the expense of  certification~\cite{ZhangAD20,ebrahimi2018hotflip}.

In this paper, we ask:
Can we develop a \emph{certified} defense to \emph{arbitrary string transformations} that applies to \emph{recursive neural networks}?

\paragraph{Our approach}
Certifying robustness involves proving that
a network's prediction is the same no matter how a given input string is perturbed. 
We assume that the \emph{perturbation space} is defined as a program
describing a set of possible  string transformations~\cite{ZhangAD20}---e.g., \emph{if you see the word ``movie'', replace it with ``film'' or ``movies''}.
Such transformations can succinctly define a  perturbation space that is exponentially large in the length of the input;
so, certification by enumerating the perturbation space is generally impractical. 

We present  \toolname{} (Abstract Recursive Certification),
an approach for \textit{certifying} robustness to programmatically defined perturbation spaces. \toolname{} can be used within an adversarial training loop
to \textit{train robust models}.
We illustrate the key ideas behind \toolname{} through a simple example.
Consider the (partial) input sentence \emph{to the movie...},
and say we are using an LSTM for prediction.
Say we have two string transformations:
(\textbf{T1}) If you see the word \emph{movie}, you can replace it with \emph{film} or \emph{movies}.
(\textbf{T2}) If you see the word \emph{the} or \emph{to}, you can delete it.
ARC avoids enumerating the large perturbation space (\cref{fig:approachfig}(a))
using two key insights.

\emph{Memoization}:
 \toolname{} exploits the recursive structure of LSTM networks, and their extensions (BiLSTMs, TreeLSTMs),
to avoid recomputing intermediate hidden states.
\toolname{}  \emph{memoizes} hidden states of prefixes shared across
multiple sequences in the perturbation space.
For example, the two sentences 
\emph{to the movie...} and \emph{to the film...}.
share the prefix \emph{to the},
and therefore we memoize the hidden state
after the word \emph{the}, as illustrated in \cref{fig:approachfig}(b) with dashed blue lines.
% We carefully derive a memoizing set of equations
% that captures the structure of programmatically
% defined string transformations.
The critical challenge is characterizing 
which strings share prefixes without having to explicitly explore the perturbation space.

% \loris{sentences below can go IMHO}
% So, instead of evaluating the LSTM from scratch on the second sentence, we can start the evaluation from the hidden state after \emph{the}, and just evaluate the LSTM on \emph{film...}.
% This form of \emph{dynamic programming} \yh{not sure if we still call this DP} is illustrated in \cref{fig:approachfig}(b).

\emph{Abstraction}:
 \toolname{} uses \emph{abstract interpretation}~\cite{DBLP:conf/popl/CousotC77} to symbolically 
represent sets of perturbed strings,
avoiding a combinatorial explosion.
Specifically, \toolname{} represents a set of strings as a hyperrectangle
in a $\mathbb{R}^n$
and propagates the hyperrectangle through the network
using interval arithmetic~\cite{ai2, DBLP:conf/iccv/GowalDSBQUAMK19}.%---each dimension of a hyperrectangle is an interval of possible values.
This idea is illustrated in \cref{fig:approachfig}(b),
where the words \emph{film} and \emph{movies} are represented as a hyperrectangle.
% Note how the words \emph{film} and \emph{movies} are represented as a single hyperrectangle, using the abstraction function $\alpha$.
% \yh{not greedily though. we merge branches that are in the same tight perturbation space.}
By \emph{joining} hidden states of different sentences (a common idea in program analysis), 
\toolname{} can perform certification efficiently.

Memoization and abstraction enable
ARC to efficiently certify robustness to
very large perturbation spaces.
% \aws{move Xu et comparison here, saying we subsume it}
Note that \toolname{} subsumes \citet{DBLP:conf/nips/XuS0WCHKLH20} because \toolname{} can certify arbitrary string transformations, while \citet{DBLP:conf/nips/XuS0WCHKLH20} only works on word substitutions.

% \aws{cite more ibp papers, in addition to ai2}

% \loris{add sec numbers to contrib}
\paragraph{Contributions}
We make the following contributions:
(1) We present \toolname{},
an approach   for training certifiably robust recursive neural networks.
    We demonstrate our approach on LSTMs, BiLSTMs, and TreeLSTMs.
(2) We present a novel application of abstract interpretation to 
    symbolically capture a large space of strings, defined programmatically,
    and propagate it through a recursive network (\Secref{sec:approach}).
(3) Our evaluation shows that \toolname{} can train models that are more robust to arbitrary perturbation spaces than those produced by existing techniques; \toolname{} can show high certification accuracy of the resulting models; and \toolname{} can certify robustness to attacks (transformations) that are out-of-scope for existing techniques (\Secref{sec:evaluation}). 

\section{Related Work}
\begin{table}[t]
    \centering
    \caption{ \toolname{} compared to other approaches. \citet{DBLP:conf/acl/YeGL20} provide probabilistic certificates. 
    }
 \renewcommand{\arraystretch}{1}
    \begin{small}
    \begin{tabular}{lccl}
        \toprule
         Approach & LSTM & Cert & Transf.   \\ 
        \midrule
        \small{\citet{ebrahimi2018hotflip}} & \cmark & \xmark & \emph{arbitrary}\\
        \small{\citet{POPQORN}} & \cmark & \cmark & $l_p$ norm\\
        \small{\citet{DBLP:conf/emnlp/HuangSWDYGDK19}} & \xmark & \cmark & substitution \\
        \small{\citet{DBLP:conf/emnlp/JiaRGL19}} & \cmark & \cmark & substitution \\
        \small{\citet{ZhangAD20}} & \xmark & \xmark & \emph{arbitrary}  \\
        \small{\citet{DBLP:conf/acl/YeGL20}} & \cmark & $\mathbb{P}$ & substitution \\
        \small{\citet{DBLP:conf/nips/XuS0WCHKLH20}} & \cmark & \cmark & substitution \\
        \citet{dong2021towards} & \cmark & \xmark & substitution \\
        \small{\toolname{} (\emph{this paper})} & \cmark & \cmark & \emph{arbitrary} \\
        \bottomrule
    \end{tabular}
    \end{small}

    \label{tab:approaches}
\end{table}

\cref{tab:approaches}
compares \toolname{} to most related approaches.

% \paragraph{Adversarial attacks in NLP} \citet{SurveyZhangSAL20} and \citet{SurveyHugTP20} conduct surveys of adversarial attacks in NLP.
% %
% There are different kinds of test-time attacks: word/character-level attacks~\cite{DBLP:conf/coling/EbrahimiLD18, DBLP:conf/milcom/PapernotMSH16, DBLP:conf/emnlp/GargR20}, 
% paraphrasing attacks~\cite{DBLP:conf/naacl/IyyerWGZ18},
% and sentence-level attacks~\cite{DBLP:conf/naacl/WangB18}.
% Some works combine different attacks~\cite{DBLP:conf/conll/BlohmJSYV18}.
% Other test-time attacks \cite{DBLP:conf/emnlp/WallaceFKGS19} look for \emph{backdoors} (universal triggers) in NLP models.

% \toolname{} is very flexible, allowing specification of attack models as sets of string transformations using \emph{match} and \emph{replace} functions
% over symbols in the strings.
% We additionally demonstrate how to extend our approach to tree transformations for TreeLSTMs~\cite{DBLP:conf/acl/TaiSM15}.

% \aws{characters and words are the same from a mathematical perspective, so not interesting to separate them. Also our approach can go beyond words, e.g., matching phrases}

\paragraph{Certification of robustness in NLP}
See \citet{SurveyLiQXL20} for a survey of robust training.
Some works focus on certifying the $l_p$ norm ball of each word embedding for LSTMs~\cite{DBLP:conf/icml/KoLWDWL19, DBLP:conf/atva/JacobyBK20} and transformers~\cite{DBLP:conf/iclr/ShiZCHH20}. 
Others focus on certifying word substitutions for CNNs~\cite{DBLP:conf/emnlp/HuangSWDYGDK19} and LSTMs~\cite{DBLP:conf/emnlp/JiaRGL19, DBLP:conf/nips/XuS0WCHKLH20}, and word deletions for the decomposable attention model~\cite{DBLP:conf/iclr/WelblHSGDSK20}. 
Existing techniques rely on abstract interpretation, such as IBP~\cite{DBLP:conf/iccv/GowalDSBQUAMK19} and CROWN~\cite{DBLP:conf/nips/ZhangWCHD18}.
We focus on certifying the robustness of LSTM models (including TreeLSTMs) to a programmable perturbation space, which is out-of-scope for existing techniques.
Note that \citet{DBLP:conf/nips/XuS0WCHKLH20} also uses memoization and abstraction to certify, but \toolname{} subsumes \citet{DBLP:conf/nips/XuS0WCHKLH20} because \toolname{} can certify arbitrary string transformations. %\yh{add comparison Xu et al. here.}
We use IBP, but our approach can use other abstract domains, such as zonotopes~\cite{ai2}.

% Randomized smoothing~\cite{RSDBLP:conf/icml/CohenRK19, RSDBLP:conf/nips/SalmanLRZZBY19} gives a probabilistic certificate of robustness.
SAFER~\cite{DBLP:conf/acl/YeGL20} is a model-agnostic approach 
that uses \emph{randomized smoothing}~\cite{RSDBLP:conf/icml/CohenRK19} to give probabilistic certificates of robustness to word substitution.
Our approach gives a non-probabilistic certificate and can handle arbitrary perturbation spaces beyond substitution.

\paragraph{Robustness techniques in NLP}
Adversarial training is an empirical defense method that can improve the robustness of models by solving a \emph{robust-optimization} problem~\cite{madry2017towards},
which minimizes worst-case (adversarial) loss.
Some techniques in NLP
use adversarial attacks to compute a lower bound on the worst-case loss~\cite{ebrahimi2018hotflip, DBLP:conf/naacl/MichelLNP19}.
ASCC~\cite{dong2021towards} overapproximates the word substitution attack space by a convex hull where a lower bound on the worst-case loss is computed using gradients.
Other techniques 
compute upper bounds on adversarial loss using abstract interpretation~\cite{DBLP:conf/iccv/GowalDSBQUAMK19, DBLP:conf/icml/MirmanGV18}. 
\citet{DBLP:conf/emnlp/HuangSWDYGDK19}
and \citet{DBLP:conf/emnlp/JiaRGL19, DBLP:conf/nips/XuS0WCHKLH20} used abstract interpretation to train CNN and LSTM models against word substitutions.
A3T~\cite{ZhangAD20} trains robust CNN models against a programmable perturbation space by combining 
adversarial training and abstraction.
Our approach uses abstract interpretation to train robust LSTMs against programmable perturbation spaces
as defined in \citet{ZhangAD20}.

% Preprocessing-based techniques improve robustness by using spelling-mistake detectors~\cite{DBLP:conf/acl/PruthiDL19, DBLP:conf/aaai/SakaguchiDPD17}.

\section{Robustness Problem and Preliminaries}
% In this section,
% we formalize the robustness-certification problem
% and provide necessary background.

% \subsection{The Robustness Problem}
We consider a classification setting with a neural network $F_\theta$ with parameters $\theta$, trained on samples from domain $\gX$ and labels from $\gY$.
The 
 domain $\gX$ is a set of  strings over a finite set of \emph{symbols} $\alphabet$ (e.g., English words or characters), i.e., $\gX = \alphabet^*$.
We use $\bfx \in \alphabet^*$ to denote a string; $x_i \in \alphabet$ to denote the $i$th element of the string; $\substring{\bfx}{i}{j}$ to denote the substring $x_i, \ldots, x_{j}$; $\epsilon$ to denote the empty string; and $\len{\bfx}$ to denote the length of the string.

% \loris{I find it awkward that we use $x_i$ instead of $\bfx_i$}
% \aws{that's what we did last time,
% so we can subscript vectors}
%We will also use $\substring{\bfx}{}{j}$ to denote the prefix ending at $x_{j-1}$ and $\substring{\bfx}{i}{}$ to denote the suffix starting at $x_i$.

\paragraph{Robustness to string transformations} A \emph{perturbation space} $S$ is a function in $\alphabet^* \to 2^{\alphabet^*}$, i.e., $S$ takes a string $\bfx$ and returns a set of possible perturbed strings obtained by modifying $\bfx$.
Intuitively, $S(\bfx)$ denotes a set of strings
that are semantically \emph{similar} to $\bfx$ and therefore
should receive the same prediction.
We assume $\bfx \in S(\bfx)$.

Given string $\bfx$ with label $y$
and a perturbation space $S$,
We say that a neural network $F_\theta$ is \textit{$S$-robust} on $(\bfx,y)$ iff
\begin{align}
    \forall \bfz \in S(\bfx) \ldotp F_\theta(\bfz) = y \label{eq:robustness}
\end{align}

Our primary goal in this paper is  to \emph{certify}, or prove, $S$-robustness (\Eqref{eq:robustness}) of the neural network for a pair $(\bfx,y)$.
Given a certification approach, we can then use it within an adversarial training loop to yield certifiably robust networks.

\paragraph{Robustness certification}
We will  certify $S$-robustness by solving
an \emph{adversarial loss} objective: 
\begin{align} 
\max_{\bfz \in S(\bfx)} L_\theta(\bfz,y)\label{eq:advloss}
\end{align} 
where we assume that the loss function $L_\theta$ is $< 0$ when $F_\theta(\bfz) = y$
and $\geq 0$ when $F_\theta(\bfz) \neq y$.
Therefore, if we can show that the solution to the above problem is $< 0$, then we have a certificate of $S$-robustness.

% The key challenge is solving the above optimization objective.
% In order to construct an $S$-robustness certificate,
% we need to compute an upper bound on the adversarial loss.

\paragraph{Certified training}
If we have a procedure to compute adversarial loss,
we can use it for \emph{adversarial training} by solving the following
\emph{robust optimization} objective~\cite{madry2017towards},
where $\mathcal{D}$ is the data distribution:
\begin{align} 
\argmin_\theta \mathop{\mathbb{E}}_{(\bfx,y) \sim \mathcal{D}} \left[\max_{\bfz \in S(\bfx)} L_\theta(\bfz,y)\right]\label{eq:advlosstrain}
\end{align} 

% \aws{I removed the challenges part}
% \paragraph{Certification challenges}
% To solve adversarial loss optimally,
% we can enumerate all strings $\bfz$ in the set $S(\bfx)$ and compute the network prediction of $\bfz$. 
% In general, it is infeasible to consider the entire perturbation space, as it can be exponentially large in the size of the input.
% Further, even if the space is practically enumerable, e.g., millions of strings, performing enumeration during training makes training impractically slow. 

% Recent works have been proposed to certify the robustness of neural networks by abstract interpretation~\yh{cite}. These works either target at the continuous $L_p$ norm region in the word embedding space or target at a specific transformation, word substitutions. 
% However, we are interested in a discrete perturbation space where there are general string transformations specified by users.

\subsection{Programmable Perturbation Spaces}
\label{sec:pps}
In our problem definition, we assumed an arbitrary perturbation 
space $S$.
We adopt the recently 
proposed specification language~\citep{ZhangAD20}
to define $S$ \emph{programmatically}
as a set of string transformations.
The language is very flexible, allowing 
the definition of a rich class of transformations
as \emph{match} and \emph{replace} functions.

\paragraph{Single transformations}
% \loris{I added the parameters $s$ and $t$ to $T$. It's ugly but maybe more precise.}
A string transformation $T$ is a pair $(\varphi, f)$, where 
$\varphi: \alphabet^s \to \{0,1\}$ is the \emph{match} function, a Boolean function that specifies the substrings (of length $s$)  to which the transformation can be applied; and
$f: \alphabet^s \to 2^{\alphabet^t}$ is the \emph{replace} function, which specifies
how the substrings matched by $\varphi$ can be replaced (with strings of length $t$).
We will call $s$ and $t$ the size of the \emph{domain}
and \emph{range} of transformation $T$, respectively.

\begin{example}
\label{exp:transformation}
In all examples, the set of symbols $\Sigma$ is English words.
So, strings are English sentences.
Let $T_\emph{del}$ be a string transformation that deletes the stop words ``\emph{to}'' and ``\emph{the}''. 
Formally, $T_\emph{del}=(\varphi_\emph{del},f_\emph{del})$, where $\varphi_\emph{del}: \alphabet^1 \mapsto \{0,1\}$ and $f_\emph{del}: \alphabet^1 \to 2^{\alphabet^0}$ are:
\vskip -0.1in
\begin{small}
\begin{align*}
    \varphi_\emph{del}(x) = \begin{cases}
    1,& x\in \{\text{``to'', ``the''}\} \\
    0,& \text{otherwise}
\end{cases}, \quad f_\emph{del}(x) = \{\epsilon\},
\end{align*}
\end{small}%

Let $T_\emph{sub}$ be a  transformation substituting the word ``\emph{movie}'' with ``\emph{movies}'' or ``\emph{film}''. Formally,
$T_\emph{sub}=(\varphi_\emph{sub},f_\emph{sub})$, where $\varphi_\emph{sub}: \alphabet^1 \mapsto \{0,1\}$ and $f_\emph{sub}: \alphabet^1 \to 2^{\alphabet^1}$ are:
\vskip -0.1in
\begin{small}
\begin{align*}
    \varphi_\emph{sub}(x) = \begin{cases}
    1,& x = \emph{``movie''}\\
    0,& \text{otherwise}
\end{cases}, f_\emph{sub}(x) = \left\{\!\begin{aligned}
&~~\emph{``film''},\\
&\emph{``movies''}
\end{aligned}\right\}
\end{align*}
\end{small}%
\end{example}
% \aws{eq outside line}

\paragraph{Defining perturbation spaces}
We can compose different string transformation to construct perturbation space $S$:
\begin{align}
    \spec=\{(T_1, \delta_1), \dots ,(T_n, \delta_n)\}, \label{eq:progspec}
\end{align}
where each $T_i$ denotes a string transformation that can be applied \emph{up to} $\delta_i \in \mathbb{N}$ times. 
% \yh{added this sentence, please check}
Note that the transformations can be applied whenever they match a \emph{non-overlapping} set of substrings and are then transformed in parallel.
We illustrate with an example
and refer to \citet{ZhangAD20} for formal semantics.

\begin{example}
\label{exp: perturbspace}
Let $\spec =\{(T_\emph{del},1),(T_\emph{sub},1)\}$ be a perturbation space that applies $T_\emph{del}$ and $T_\emph{sub}$ to the given input sequence up to once each.
If $\bfx=$``\emph{to the movie}'', a subset of the perturbation space $\spec(\bfx)$ is shown in \cref{fig:approachfig}(a). 
% \yh{In intro, we only allow to delete ``the'', need we unify this later?}
\end{example}

\paragraph{Decomposition}
$S=\{(T_i, \delta_i)\}_i$ can be decomposed into $\prod (\delta_i + 1)$ subset perturbation spaces by considering all smaller combinations of $\delta_i$.
We denote the decomposition of perturbation space $S$ as $\tightperturbset{S}$, and exemplify below:

\begin{example}
\label{ex:decs}
$S_1=\{(T_\emph{del},2)\}$ can be decomposed to a set of three perturbation spaces $\tightperturbset{S_1} = 
\{\varnothing,  \{(T_\emph{del},1)\}, \{(T_\emph{del},2)\}\}$,
while 
 $S_2=\{(T_\emph{del},1),(T_\emph{sub},1)\}$ can be decomposed to a set of four perturbation spaces
 \vskip -0.1in
 \begin{small}
 \[
     \tightperturbset{S_2}=\{\varnothing, \{(T_\emph{del},1)\}, \{(T_\emph{sub},1)\}, \{(T_\emph{del},1),(T_\emph{sub},1)\}\}\]
 \end{small}%
where $\varnothing$ is the perturbation space  with no transformations, i.e.,
if $S = \varnothing$, then $S(\bfx) = \{\bfx\}$
for any string $\bfx$.
\end{example}

% \yh{move to here}
We use notation $S_{k\downarrow{}}$ to denote $S$
after reducing $\delta_k$ by 1;
therefore, $S_{k\downarrow{}} \in \tightperturbset{S}$.

% \begin{example}
% The definition of $\sqcup$ is consistent when $a,b$ are two tensors of intervals. Suppose $\lb{a}=(0,0)^\top, \ub{a}=(0,1)^\top, \lb{b}=(-1,2)^\top, \ub{b}=(0,3)^\top$, the lower bound and upper bound of $a \sqcup b$ are $(-1, 0)^\top$ and $(0,3)^\top$.
% \end{example}

\subsection{The LSTM Cell}
We  focus our exposition on LSTMs.
%
% An LSTM takes as input a string $\bfx$ and returns a prediction;
% this is done by recursively applying an LSTM \emph{cell} to each symbol $x_i$ of $\bfx$.
%
% We assume that our alphabet is that of \emph{words} (i.e., word-level modeling, see Example~\ref{exp:transformation}).
% and that each word is given a real-valued vector
% generated, for example, using word embeddings.
% Thus, we use the $i$th word of string $x_i$ as a vector.
% \aws{maybe early on in prob def we just say our alphabet is a dictionary $\Sigma$}
%
An LSTM cell is a function, denoted $\lstm$, that takes as input a symbol $x_i$ and the previous \emph{hidden state} and \emph{cell state}, and outputs the hidden state and the cell state at the current time step.
For simplicity, we use $\hstate_i$ to denote the concatenation of the hidden and cell states at the $i$th time step,
and simply refer to it as the \emph{state}.
Given string $\bfx$, we define $\hstate_i$ as follows:
\begin{align*}
    \hstate_i &= \lstm(x_i, \hstate_{i-1}) \quad \hstate_i, \hstate_{i-1}\in \mathbb{R}^{d}, %\label{eq:pointformlstm2}
\end{align*}
where     $\hstate_0 = 0^{d}$
and $d$ is the dimensionality of the state.
For a string $\bfx$ of length $n$,
we say that $h_n$ is the \emph{final state}.

% It will be useful for us to apply LSTMs to substrings
% of a string $\bfx$.
For a string $\bfx$ and state $h$,
We use $\lstm(\bfx,h)$ to denote 
\[\lstm(x_{\len{\bfx}}, \lstm(\ldots, \lstm(x_1 ,\hstate)\ldots))\]
E.g., $\lstm(\bfx, \hstate_0)$ is the final state of the LSTM applied to  $\bfx$.

\section{\toolname{}: Abstract Recursive Certification}
\label{sec:approach}
In this section, we present our technique for proving $S$-robustness of an LSTM
on $(\bfx,y)$.
Formally, we do this by computing the adversarial loss,
$\max_{\bfz \in S(\bfx)} L_\theta(\bfz,y)$.
Recall that
if the solution is $< 0$, then we have proven $S$-robustness.
To solve adversarial loss optimally, we effectively need to evaluate the LSTM on all of $S(\bfx)$ and collect all final states:
\begin{align}
F = \{\lstm(\bfz,h_0) \mid \bfz \in S(\bfx)\} \label{eq: theF}
\end{align}
Computing $F$  precisely is challenging,
as $S(\bfx)$ may be prohibitively large.
Therefore, we propose to compute a superset
of $F$, which we will call $\ibp{F}$.
This superset will therefore yield an upper bound on the adversarial loss. 
We prove $S$-robustness if the upper bound is $<0$.

To compute $\ibp{F}$, we present two key ideas
that go hand-in-hand:
 In \Secref{sec:concrete}, we observe that strings in the perturbation space share common prefixes,
and therefore we can \emph{memoize} hidden states to reduce the number of evaluations of LSTM cells---a form of dynamic programming.
We carefully derive the set of final states $F$
as a system of memoizing equations.
The challenge of this derivation is characterizing which strings share common prefixes without explicitly exploring the perturbation space.
In \Secref{sec:abstract},
we apply abstract interpretation
to efficiently and soundly solve the system of memoizing equations,
thus computing an overapproximation 
$\ibp{F} \supseteq F$.

% These  ideas result in an efficient approach 
% for computing $\ibp{F}$.
% that is linear\yh{polynomial?} in the length of the input string
% and the radius of the perturbation space (defined by $\delta_i$s).

% \input{sections/ablation}
% \begin{figure*}[t]
%     \centering
%     \includegraphics[width=1.8\columnwidth]{imgs/illu_state_new.png}
%     \caption{Illustration of two cases of \Eqref{eq:contransition1}}
% \end{figure*}

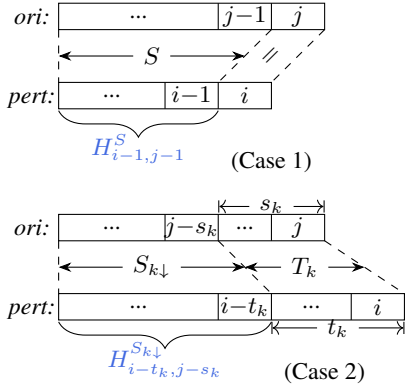
\begin{figure}[t]
    \centering
    \begin{tikzpicture}[scale=.7]
% 	\draw[help lines,step=5mm,gray!20] (0,0) grid (7,-5);
\tikzstyle{every node}=[font=\small]
    \pgfmathsetmacro{\y}{1.5}
    \pgfmathsetmacro{\width}{0.5}
    \pgfmathsetmacro{\la}{5}
    \pgfmathsetmacro{\lb}{4}
    \pgfmathsetmacro{\lc}{6.5}
    \begin{scope}[on grid]
        \node (original) at (0,-0.25) {\emph{ori:}};
        \node[below=1.1 of original] (perturbed) {\emph{pert:}};
        \draw (0.5, 0) -- ++(\la, 0) -- ++(0,-\width) -- ++ (-\la, 0)-- cycle;
        \draw (0.5, -\y) -- ++(\lb, 0) -- ++(0,-\width) -- ++ (-\lb, 0)-- cycle;
        \node at (\la, -0.25) {$j$};
        \node at (\la-1, -0.25) {$j{-}1$};
        \node at (2, -0.25) {...};
        \node at (\lb, -0.25-\y) {$i$};
        \node at (\lb-1, -0.25-\y) {$i{-}1$};
        \node at (1.5, -0.25-\y) {...};
        \draw (\la - 0.5, 0) -- ++(0, -\width);
        \draw (\la - 1.5, 0) -- ++(0, -\width);
        \draw (\lb - 0.5, -\y) -- ++(0, -\width);
        \draw (\lb - 1.5, -\y) -- ++(0, -\width);
        \draw[dashed] (\la+0.5, -\width) -- (\lb+0.5, -\y);
        \draw[dashed] (\la-0.5, -\width) -- (\lb-0.5, -\y);
        \draw[dashed] (0.5, -\width) -- (0.5, -\y);
        \node (S) at (2.25, -1) {$S$};
        \draw[-{Stealth[slant=0]}] (S) -- (0.5, -1);
        \draw[-{Stealth[slant=0]}] (S) -- (4, -1);
        \node[rotate=45] at (4.5, -1) {$=$};
        \draw [decorate,decoration={brace,amplitude=10pt,mirror},xshift=0pt,yshift=-1pt]
        (0.5,-\width-\y) -- ++(\lb - 1,0) node [RoyalBlue,midway,xshift=0cm, yshift=-0.5cm] 
        {\footnotesize $H_{i-1,j-1}^S$};
        \node at (4.5, -3) {(Case 1)};
    \end{scope}
    
    \begin{scope}[on grid, yshift=-4cm]
        \node (original) at (0,-0.25) {\emph{ori:}};
        \node[below=1.1 of original] (perturbed) {\emph{pert:}};
        \draw (0.5, 0) -- ++(\la, 0) -- ++(0,-\width) -- ++ (-\la, 0)-- cycle;
        \draw (0.5, -\y) -- ++(\lc, 0) -- ++(0,-\width) -- ++ (-\lc, 0)-- cycle;
        
        \node at (\la, -0.25) {$j$};
        \node at (\la-1, -0.25) {...};
        \node at (\la-2, -0.25) {$j{-}s_k$};
        \node at (1.5, -0.25) {...};
        \node (sk) at (\la-0.5, 0.15) {\small $s_k$};
        \draw[->|] (sk) -- (\la-1.5, 0.15);
        \draw[->|] (sk) -- (\la+0.5, 0.15);
        
        \node at (\lc, -0.25-\y) {$i$};
        \node at (\lc-1.25, -0.25-\y) {...};
        \node at (\lc-2.5, -0.25-\y) {$i{-}t_k$};
        \node at (2, -0.25-\y) {...};
        \node (tk) at (\lc-0.75, -\y-0.15-\width) {\small $t_k$};
        \draw[->|] (tk) -- (\lc-2, -\y-0.15-\width);
        \draw[->|] (tk) -- (\lc+0.5, -\y-0.15-\width);

        \draw (\la - 0.5, 0) -- ++(0, -\width);
        \draw (\la - 1.5, 0) -- ++(0, -\width);
        \draw (\la - 2.5, 0) -- ++(0, -\width);
        \draw (\lc - 0.5, -\y) -- ++(0, -\width);
        \draw (\lc - 2, -\y) -- ++(0, -\width);
        \draw (\lc - 3, -\y) -- ++(0, -\width);
        \draw[dashed] (\la+0.5, -\width) -- (\lc+0.5, -\y);
        \draw[dashed] (\la-1.5, -\width) -- (\lc-2, -\y);
        \draw[dashed] (0.5, -\width) -- (0.5, -\y);
        
        \node (Sk) at (2.25, -1) {$S_{k\downarrow{}}$};
        \draw[-{Stealth[slant=0]}] (Sk) -- (0.5, -1);
        \draw[-{Stealth[slant=0]}] (Sk) -- (4, -1);
        
        \node (Tk) at (5.125, -1) {$T_k$};
        \draw[-{Stealth[slant=0]}] (Tk) -- (4, -1);
        \draw[-{Stealth[slant=0]}] (Tk) -- (6.25, -1);
        
        \draw [decorate,decoration={brace,amplitude=10pt,mirror},xshift=0pt,yshift=-1pt]
        (0.5,-\width-\y) -- ++(\lc - 2.5,0) node [RoyalBlue,midway,xshift=0cm, yshift=-0.5cm] 
        {\footnotesize $H_{i{-}t_k,j{-}s_k}^{S_{k\downarrow{}}}$};
        
        \node at (5.5, -3) {(Case 2)};
    \end{scope}
\end{tikzpicture}
    \caption{Illustration of two cases of \Eqref{eq:contransition1}.}
    \label{fig:illu_state_new}
\end{figure}

\subsection{Memoizing Equations of Final States}
\label{sec:concrete}
% We now derive a set of
% memoizing equations defining $F$.
% We begin by showing how to characterize sets of
% intermediate states of an LSTM with respect to a perturbation space.

% \subsubsection{Defining prefix states}
\paragraph{Tight Perturbation Space} 
Given a perturbation space $S$, we shall use $\tightperturb{}$ 
to denote the \emph{tight} perturbation space where each transformation $T_j$ in $S$ is be applied \textbf{exactly} $\delta_j$ times. 

Think of the set of all strings
in a perturbation space as a tree, like in \cref{fig:approachfig}(b),
where strings that share prefixes
share LSTM states.
We want to characterize a subset $H_{i,j}^S$ of LSTM
states at the $i$th layer where the perturbed prefixes have had \emph{all} transformations in a space $S$ applied on the original prefix $\substring{\bfx}{1}{j}$.

We formally define  $H_{i,j}^S$ as follows:
\vskip -0.1in
\begin{small}
\begin{align}
    H_{i,j}^S = \{\lstm(\bfz, \hstate_0)\mid \bfz \in \tightperturb{}(\substring{\bfx}{1}{j}) \wedge \len{\bfz}=i\} \label{eq: complexstate}
\end{align}
\end{small}%
By definition, the base case $H_{0,0}^\varnothing = \{0^d\}$.

% \yh{only the following example uses the $T_\emph{del}$ by deleting `the` and `to`.}
\begin{example}
\label{ex: state}
Let $\bfx =$ ``\emph{to the movie}''.
Then, $H_{1,2}^{\{(T_\emph{del}, 1)\}} =\{\lstm(\emph{``the''},h_0), \lstm(\emph{``to''},h_0)\}$. 
These states result from deleting the first and second words of the prefix ``\emph{to the}'',
respectively.
We also have
$H_{2,2}^\varnothing = \{\lstm(\emph{``to the''}, h_0)\}$.
\end{example}

% The set of final states of strings in 
%  $\tightperturb{(\bfx)}$ are
% \begin{align}
%   \bigcup_{0 \le i \le \maxlen{\bfx}} H_{i,\len{\bfx}}^S \label{eq: finalstateoftight}
% \end{align}
% where $\maxlen{\bfx}$ is the upper bound of the length of the perturbed strings. $\maxlen{\bfx}$ can be overapproximated by
% \[\maxlen{\bfx} = \len{\bfx} + \sum_{(T_k, \delta_k) \in S} \max(t_k {-} s_k, 0)\delta_k\]
% where $f_k: \alphabet^{s_k} \to 2^{\alphabet^{t_k}} $.
% Because $t_k,s_k,\delta_k$ are small constants, we can regard $\maxlen{\bfx}$ as a term that is linear in the length of the original string $\len{\bfx}$, i.e., $\maxlen{\bfx} = O(\len{\bfx})$.

The set of final states of strings in 
 $\tightperturb{(\bfx)}$ is
\begin{align}
\bigcup_{i \ge 0} H_{i,\len{\bfx}}^S.
   \label{eq: finalstateoftight}
\end{align}

\paragraph{Memoizing equation}
We now demonstrate how to rewrite \Eqref{eq: complexstate} by explicitly applying the transformations defining the perturbation space $S$.
Notice that each $H_{i,j}^S$ comes from two sets of strings: (1) strings whose suffix (the last character) is not perturbed by any transformations (the first line of \Eqref{eq:contransition1}), and (2) strings whose suffix is perturbed by $T_k = (\varphi_k,f_k)$ (the second line of \Eqref{eq:contransition1}), as illustrated in \Figref{fig:illu_state_new}. 
Thus, we derive the final equation and then immediately show an example:
\vskip -0.1in
\begin{small}
\begin{align}
%   H_{i,j}^S = & \bigcup_{\substack{1\leq k \leq |S|\\ \varphi_k(\substring{\bfx}{a}{b})=1}} \{\lstm(\bfz, h) \mid  \bfz \in f_k(\substring{\bfx}{a}{b}), h \in H_{i-t_k,j-s_k}^{S_{k\downarrow{}}}\}\nonumber \\
%   &~~~~~~ \cup\ \{\lstm(x_{j}, h) \mid h \in H_{i-1,j-1}^S\} \label{eq:contransition1}
 &H_{i,j}^S = \{\lstm(x_{j}, h) \mid h \in H_{i{-}1,j{-}1}^S\} \cup \nonumber\\
 &\bigcup_{\substack{1\leq k \leq |S|\\ \varphi_k(\substring{\bfx}{a}{b})=1}} \{\lstm(\bfz, h) \mid  \bfz \in f_k(\substring{\bfx}{a}{b}), h \in H_{i{-}t_k,j{-}s_k}^{S_{k\downarrow{}}}\}\label{eq:contransition1}
\end{align}
\end{small}%
where 
$a {=} j {-} s_k + 1$ and
$b {=} j$.

We compute \Eqref{eq:contransition1} in a bottom-up fashion, starting from $H_{0,0}^\varnothing=\{0^d\}$ and increasing $i,j$ and considering every possible perturbation space in the decomposition of $S$, $\tightperturbset{S}$.

\begin{lemma}
\label{lemma:4.2}
 \Eqref{eq:contransition1} and \Eqref{eq: complexstate} are equivalent.
\end{lemma}

\begin{example}
% \yh{todo: example to show Eq 10.}
Consider computing $H_{1,2}^{\{(T_\emph{del}, 1)\}}$. 
We demonstrate how to derive states from \Eqref{eq:contransition1}:
\vskip -0.1in
\begin{small}
\begin{align}
%   H_{1,2}^{\{(T_\emph{del}, 1)\}} = & \{\lstm(\bfz, h) \mid  \bfz \in f_\emph{del}(\text{``the''}), h \in H_{1,1}^\varnothing\} \cup \nonumber \\
%   & \{\lstm(\text{``the''}, h) \mid h \in H_{0,1}^{\{(T_\emph{del}, 1)\}}\} \label{eq:concreteex}
H_{1,2}^{\{(T_\emph{del}, 1)\}} =&  \{\lstm(\text{``the''}, h) \mid h \in H_{0,1}^{\{(T_\emph{del}, 1)\}}\} \cup \nonumber \\
  & \{\lstm(\bfz, h) \mid  \bfz \in f_\emph{del}(\text{``the''}), h \in H_{1,1}^\varnothing\} \label{eq:concreteex}
\end{align}
\end{small}%
Assume $H_{1,1}^\varnothing=\{\lstm(\text{``to''}, h_0)\}$ and 
$H_{0,1}^{\{(T_\emph{del}, 1)\}}=\{h_0\}$ are computed in advance. 
The first line of \Eqref{eq:concreteex} evaluates to $\{\lstm(\text{``the''}, h_0)\}$, which corresponds to deleting the first word of the prefix ``to the''.
Because $\bfz$ can only be an empty string, the second line of \Eqref{eq:concreteex} evaluates to $\{\lstm(\text{``to''}, h_0)\}$, which corresponds to deleting the second word of ``to the''. 
The dashed green line in \cref{fig:approachfig}(b) shows the computation of \Eqref{eq:concreteex}.
\end{example}

\paragraph{Defining Final States using Prefixes}
Finally, we compute the set of final states, $F$, by considering all perturbation spaces in the decomposition of $S$.
\begin{equation}
F = \bigcup_{S' \in \tightperturbset{S}} \bigcup_{i \ge 0} H_{i,\len{\bfx}}^{S'} \label{eq:hfinal}
\end{equation}
% \yh{please check}
% Intuitively, assuming $S$ has a single transformation, $F$ is the union of final states of strings with $0$ transformations, exactly $1$ transformation, exactly $2$ transformations, etc. 

% \aws{note that in the equation above we shouldn't use $S^=$,
% because the "tightness" is implicit in the definition of $H_i^S$. I.e., you give it an $S$ and it treats it tightly.}

\begin{theorem}
\label{theorem: theorem4.1}
\Eqref{eq:hfinal} is equivalent to \Eqref{eq: theF}.
\end{theorem}

\begin{example}
Let $\spec=\{(T_\emph{del},1),(T_\emph{sub},1)\}$ and $\bfx=$``to the movie''.
$F$ is the union of four final states, $H_{3,3}^\varnothing$ (no transformations), $H_{2,3}^{\{(T_\emph{del}, 1)\}}$ (exactly 1 deletion), $H_{3,3}^{\{(T_\emph{sub}, 1)\}}$ (exactly 1 substitution), and $H_{2,3}^{\{(T_\emph{del}, 1), (T_\emph{sub}, 1)\}} $ (exactly 1 deletion and 1 substitution).
% Other $H_{i,j}^{S'}$ are either non-final or empty.
\end{example}

\subsection{Abstract Memoizing Equations}
\label{sec:abstract}
% We have characterized
% the set of final states $F$ as a memoizing system of equations.
Memoization avoids recomputing hidden states, but it still incurs a combinatorial explosion.
We employ abstract interpretation~\cite{DBLP:conf/popl/CousotC77} to solve the equations efficiently by overapproximating the set $F$.
See \citet{nnvbook} for details on abstractly interpreting neural networks.

\paragraph{Abstract Interpretation}
\label{sec: ibp_def}
The \emph{interval domain}, or \emph{interval bound propagation},
allows us to evaluate a function %
on an infinite set of inputs
represented as a hyperrectangle in $\mathbb{R}^n$.

\textit{Interval domain.}
We define the interval domain over scalars---the extension to vectors is standard.
We will use an interval $[l,u] \subset \mathbb{R}$, where $l,u\in\mathbb{R}$ and $l \leq u$,
to denote the set of all real numbers between $l$ and $u$, inclusive.
% We call  $l$ the lower bound and $u$ the upper bound.

For a finite set $X \subset \mathbb{R}$,
the \emph{abstraction} operator 
gives the tightest interval containing $X$,
as follows: $\alpha(X) = [\min(X),\max(X)]$.
Abstraction allows us to compactly represent a large set of strings.

\begin{example}
Suppose the words $x_1 = \emph{``movie''}$ and $x_2 = \emph{``film''}$
have the 1D embedding $0.1$
and $0.15$, respectively.
Then, $\alpha(\{x_1,x_2\}) = [0.1,0.15]$.
For $n$-dimensional embeddings,
we simply compute an abstraction of every dimension,
producing a vector of intervals.
\end{example}

The \emph{join} operation, $\sqcup$,
 produces the smallest interval containing two intervals:
 $[l,u] \sqcup [l',u'] = [\min(l,l'), \max(u,u')]$.
% Unlike in program analysis, join is rarely  used in network certification, due to lack of branching control-flow.
We will use joins to merge hidden states 
resulting from different strings in the perturbation space (recall \cref{fig:approachfig}(b)).

\begin{example}
Say we have two sets of 1D LSTM states
represented as intervals,
$[1,2]$ and $[10,12]$.
Then $[1,2] \sqcup [10,12] = [1,12]$.
Note that $\sqcup$ is an overapproximation of $\cup$,
introducing elements in neither interval.
\end{example}

\textit{Interval transformers.} 
To evaluate a neural network on  intervals,
we lift neural-network operations to interval arithmetic---%
\emph{abstract transformers}.
For a function $g$, we use $\ibp{g}$ to denote
its abstract transformer.
We use the transformers proposed by \citet{ai2,DBLP:conf/emnlp/JiaRGL19}.
We illustrate transformers for addition and any monotonically increasing function $g:\mathbb{R}\to\mathbb{R}$ (e.g., ReLU, tanh).
\vskip -0.1in
\begin{small}
\begin{align*}
    [l,u] \ibp{+}[l',u'] = [l+l', u+u'],\quad
    \quad \ibp{g}([l,u]) = [g(l),g(u)] 
\end{align*}
\end{small}%
Note how, for monotonic functions $g$, the
abstract transformer $\ibp{g}$ simply applies $g$
to the lower and upper bounds.

\begin{example}
When applying to the ReLU function,
 $\widehat{\mathrm{relu}}([-1,2]) = [\mathrm{relu}(-1),\mathrm{relu}(2)] = [0,2]$.
\end{example}

An abstract transformer
$\ibp{g}$
must be \emph{sound}: for any interval $[l,u]$
and $x \in [l,u]$, we have $g(x) \in \ibp{g}([l,u])$.
We use $\ibp{\lstm}$ 
to denote an abstract transformer of an LSTM cell.
It takes an interval of symbol embeddings and an interval of  states.
We use the definition of $\ibp{\lstm}$ given by \citet{DBLP:conf/emnlp/JiaRGL19}.

% \aws{i removed soundness}

% \paragraph{Soundness}
% An abstract transformer $\ibp{g}$ of a function $g$
% is said to be sound iff it overapproximates the results of
% $g$.
%
%  $\ibp{g}$
% is \emph{sound} iff for any interval $[l,u]$
% and any $x \in [l,u]$, we have $g(x) \in \ibp{g}([l,u])$.
% %
% Soundness is critical for certification,
% as we need  capture all final states $F$.

\paragraph{Abstract Memoizing Equations}
We now show how to solve
\Eqref{eq:contransition1}
and \Eqref{eq:hfinal} 
using abstract interpretation.
We do this by rewriting the equations 
using operations over intervals.
Let $\ibp{H}_{0,0}^\varnothing = \alpha(\{0^d\})$, then
\begin{align*}
%   \ibp{H}_{i,j}^S = & \bigsqcup_{\substack{1\leq k \leq |S|\\ \varphi_k(\substring{\bfx}{a}{b})=1}} \widehat{\lstm}(\alpha(f_k(\substring{\bfx}{a}{b})), \widehat{H}_{i-t_k,j-s_k}^{S_{-k}}) \\
%   &~~~~~~\sqcup \ \widehat{\lstm}(\alpha(\{x_{j}\}), \widehat{H}_{i-1,j-1}^S)\\
  \ibp{H}_{i,j}^S =&\ \widehat{\lstm}(\alpha(\{x_{j}\}), \widehat{H}_{i{-}1,j{-}1}^S) \ \sqcup \\ 
  &\bigsqcup_{\substack{1\leq k \leq |S|\\ \varphi_k(\substring{\bfx}{a}{b})=1}} \widehat{\lstm}(\alpha(f_k(\substring{\bfx}{a}{b})), \widehat{H}_{i{-}t_k,j{-}s_k}^{S_{k\downarrow{}}}) \\
%   \label{eq:abstract}~\\
 \ibp{F} = & \bigsqcup_{S' \in \tightperturbset{S}} \bigsqcup_{i \ge 0}  \ibp{H}_{i, \len{\bfx}}^{S'} \nonumber
\end{align*}
where $a$ and $b$ are the same in \Eqref{eq:contransition1}.

The two key ideas are (1) representing sets of possible LSTM inputs abstractly as intervals, using $\alpha$;
and (2) joining intervals of 
 states, using $\sqcup$.
These two ideas ensure that we 
efficiently solve the system of equations, producing an overapproximation $\ibp{F}$.

The above abstract equations give us a compact overapproximation of $F$ that can be computed with a number of steps that is linear in the length of the input. Even though we can have $O(\len{\bfx}^2)$ number of $H_{i,j}^S$ for a given $S$, only $O(\len{\bfx})$ number of $H_{i,j}^S$ are non-empty. This property is used in \Theref{theorem: theorem4.2} and will be proved in the appendix.
% \yh{I remove $\maxlen{\bfx}$, does the above still make sense?}

% \begin{example}
% Consider $\bfx=$``to the movie''.
% If $S=\{(T_\emph{del},1)\}$, states $H_{0,1}^S, H_{1,2}^S, H_{2,3}^S$ are non-empty. If $S =\{(T_\emph{sub},1)\}$, $H_{3,3}^S$ are non-empty.
% \end{example}

% \yh{merged the following theorems}
\begin{theorem}(Soundness \& Complexity)
\label{theorem: theorem4.2}
$F \subseteq \widehat{F}$ and the
number of LSTM cell evaluations needed to compute $\ibp{F}$ is $O(\len{\bfx}\cdot n\cdot\prod_{i=1}^n \delta_i)$.
\end{theorem}

For practical perturbations spaces (see \Secref{sec:evaluation}), the quantity $n\prod_{i=1}^n \delta_i$ is typically small and can be considered constant.

% We also make the assumption that
% the range of \emph{replace} functions $f_j$ can be abstracted offline.
% For example, if $f_j(x)$ defines the set of synonyms of a word $x$, we can compute $\alpha(f_j(x))$ offline for every word in the dictionary.
% %

% \aws{I think redrawing the running at this point with $H_i$ etc. would be very instructive} 

% \subsection{Ablation of Tight Perturbation Spaces}

\paragraph{Extension to Bi-LSTMs and Tree-LSTMs}\label{ssec:ext}
A \emph{Bi-LSTM} performs a forward and a backward pass on the input.  
The forward pass is the same as the forward pass in the original LSTM. 
For the backward pass, we reverse the input string $\bfx$, the input of the match function $\varphi_i$ and 
the input/output of the replace function $f_i$ of each transformation.

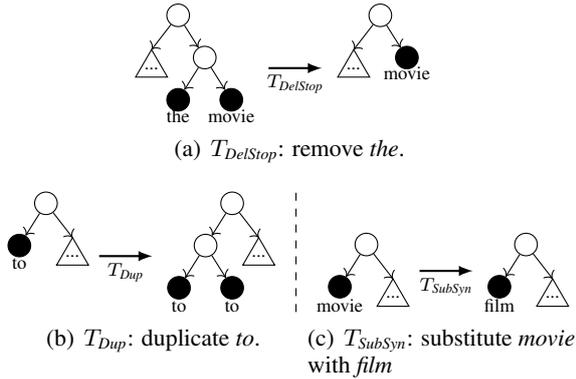
\begin{figure}[!t]
    \centering\small
    \subfigure[$\Tdelword{}$: remove \textit{the}.]{\begin{tikzpicture}[scale=0.7, mytext/.style={anchor=west, text width=110, text height=1}, nonleafnode/.style={draw, circle}, leafnode/.style={draw, circle, fill=black}, treelink/.style={->}]
    \tikzstyle{every node}=[font=\scriptsize]
    % \draw[help lines,step=5mm,gray!20] (0,0) grid (5.5,-2.5);
    \node[nonleafnode] (A1) at (0.8, 0) {}; % 5.5 * 40% / 2
    \node[nonleafnode] (A2) at (4.6, 0) {}; % 5.5 * (1 - 35% / 2)
    \node[nonleafnode] (B1) at (1.3, -0.8) {}; % 5.5 * 40% * 75%
    \node[leafnode] (C1) at (0.8, -1.6) {};
    \node[leafnode] (D1) at (1.8, -1.6) {}; % 5.5 * 40%
    \node[leafnode] (C2) at (5.1, -0.8) {};
    
    \draw (0.3,-0.65) -- ++(0.3,-0.52) -- ++(-0.6,0) -- cycle;
    \draw (4.1,-0.65) -- ++(0.3,-0.52) -- ++(-0.6,0) -- cycle; % 5.5 * (1 - 35% * 75%) 
    
    \node at (0.3, -1) {...};
    \node at (4.1,-1) {...};
    \node at (0.8, -1.9) {the};
    \node at (1.8, -1.9) {movie};
    \node at (5.1, -1.1) {movie};
    \node at (3, -1.3) {$\Tdelword{}$}; % 5.5 * 55%
    
    \draw[treelink] (A1) -- (0.3, -0.65);
    \draw[treelink] (A1) -- (B1);
    \draw[treelink] (B1) -- (C1);
    \draw[treelink] (B1) -- (D1);
    \draw[treelink] (A2) -- (4.1, -0.65);
    \draw[treelink] (A2) -- (C2);
    \draw[>=latex, ->, thick] (2.5, -1) -- ++(1, 0);
    % \draw[dashed] (6,0.2) -- ++(0, -2.3);
\end{tikzpicture}\label{fig:transdelstop}}
    \subfigure[$\Tinsword{}$: duplicate \textit{to}.]{\begin{tikzpicture}[scale=0.7, mytext/.style={anchor=west, text width=110, text height=1}, nonleafnode/.style={draw, circle}, leafnode/.style={draw, circle, fill=black}, treelink/.style={->}]
    \tikzstyle{every node}=[font=\scriptsize]
    % \draw[help lines,step=5mm,gray!20] (0,0) grid (5.5,-2);
    \node[nonleafnode] (A) at (0.5, 0) {};
    \node[leafnode] (B) at (0, -0.8) {};
    
    \draw (1, -0.65) -- ++(0.3,-0.52) -- ++(-0.6,0) -- cycle;
    
    \node at (0, -1.15) {to}; 
    \node at (1, -1) {...};
    \node at (2, -1.3) {$\Tinsword{}$};
    
    \draw[treelink] (A) -- (B);
    \draw[treelink] (A) -- (1, -0.65);
    \draw[>=latex, ->, thick] (1.5, -1) -- ++(1, 0);
    
    \begin{scope}[xshift=3.5cm]
        \node[nonleafnode] (A1) at (0.5, 0) {};
        \node[nonleafnode] (B1) at (0, -0.8) {};
        \node[leafnode] (C1) at (-0.5, -1.6) {};
        \node[leafnode] (D1) at (0.5, -1.6) {};
        
        \draw (1, -0.65) -- ++(0.3,-0.52) -- ++(-0.6,0) -- cycle;
        
        \node at (-0.5, -1.95) {to}; 
        \node at (0.5, -1.95) {to}; 
        \node at (1, -1) {...};
        
        \draw[treelink] (A1) -- (B1);
        \draw[treelink] (A1) -- (1, -0.65);
        \draw[treelink] (B1) -- (C1);
        \draw[treelink] (B1) -- (D1);
	\end{scope}
	\draw[dashed] (5.2,0.2) -- ++(0, -2.3);
\end{tikzpicture}\label{fig:transdup}}
    \subfigure[$\Tsubword{}$: substitute \textit{movie} with \textit{film}]{% \trimbox{-1cm 0cm -1cm 0cm}{ 

\begin{tikzpicture}[scale=0.7, mytext/.style={anchor=west, text width=110, text height=1}, nonleafnode/.style={draw, circle}, leafnode/.style={draw, circle, fill=black}, treelink/.style={->}]
    \tikzstyle{every node}=[font=\scriptsize]
    % \draw[help lines,step=5mm,gray!20] (0,0) grid (5.5,-2);
    \node[nonleafnode] (A) at (0.5, 0) {};
    \node[leafnode] (B) at (0, -0.8) {};
    
    \draw (1, -0.65) -- ++(0.3,-0.52) -- ++(-0.6,0) -- cycle;
    
    \node at (0, -1.15) {movie}; 
    \node at (1, -1) {...};
    \node at (2, -0.8) {$\Tsubword{}$};
    
    \draw[treelink] (A) -- (B);
    \draw[treelink] (A) -- (1, -0.65);
    \draw[>=latex, ->, thick] (1.5, -0.5) -- ++(1, 0);
    
    \begin{scope}[xshift=3cm]
        \node[nonleafnode] (A) at (0.5, 0) {};
        \node[leafnode] (B) at (0, -0.8) {};
        
        \draw (1, -0.65) -- ++(0.3,-0.52) -- ++(-0.6,0) -- cycle;
        
        \node at (0, -1.15) {film}; 
        \node at (1, -1) {...};
        
        \draw[treelink] (A) -- (B);
        \draw[treelink] (A) -- (1, -0.65);
	\end{scope}
\end{tikzpicture}

% }
\label{fig:transsubsyn}}
    \caption{Examples of tree transformations.}
    \label{fig: treetrans}
\end{figure}

A \emph{Tree-LSTM} takes trees as input. We can define the programmable perturbation space over trees in the same form of \Eqref{eq:progspec}, where $T_i$ is a tree transformation.
We show some examples of tree transformations in \Figref{fig: treetrans}.
$\Tdelword{}$ (\Figref{fig:transdelstop}) removes a leaf node with a stop word in the tree. After removing, the sibling of the removed node becomes the new parent node.
$\Tinsword{}$ (\Figref{fig:transdup}) duplicates a word in a leaf node by first removing the word and expanding the leaf node with two children, each of which contains the previous word.
$\Tsubword{}$ (\Figref{fig:transsubsyn}) substitutes a word in the leaf node with one of its synonyms. 

We provide the formalization of \toolname{} on Bi-LSTMs and Tree-LSTMs in the appendix.

\section{Evaluation}
\label{sec:evaluation}
%\footnote{anonymous link}
% \aws{add machine details to supp}
% We conduct all experiments on a server running Ubuntu 18.04.5 LTS with V100 32GB GPUs and Intel Xeon Gold 5115 CPUs running at 2.40GHz.
% In this section, we evaluate \toolname{} by answering the following research questions:
% \begin{description}
%     \item[RQ1:] Does \toolname{} improve robustness in rich perturbation spaces for LSTM, Tree-LSTM, and Bi-LSTM?
%     % \item[RQ2:] How does \toolname{} compare to baselines\loris{how is this question different than the previous one?} \yh{baselines that cannot handle general transformations or certify the robustness.}?
%     \item[RQ2:] How does the size of the perturbation space affect the effectiveness of \toolname{}?
% \end{description}

% \subsection{Experimental Setup}
We implemented \toolname{} in PyTorch.
The source code is available online\footnote{\url{https://github.com/ForeverZyh/certified_lstms}} and provided in the supplementary materials.

\paragraph{Datasets}
We use three datasets:  \emph{IMDB}~\cite{imdbdataset}, \emph{Stanford Sentiment Treebank} (SST)~\cite{sst2}, and SST2, a two-way class split of SST.
% All datasets consist of movie reviews and human annotations of their sentiment. 
IMDB and SST2 have reviews in sentence form and binary labels. 
SST has reviews in the \emph{constituency parse tree} form and five labels.

% \aws{say something that Trees and Bi are in Appendix early on}
\paragraph{Perturbation Spaces}
\begin{table}[t!]
    \centering
     \caption{String transformations.}
\renewcommand{\arraystretch}{1}
    % \setlength{\tabcolsep}{1pt}
    % \rowcolors{2}{}{gray!10}
    \begin{small}
    \begin{tabular}{ll}
        \toprule
         Trans & Description\\
        \midrule
          $\Tdelword{}$ &   \textbf{delete} a \textbf{stop word}, e.g., \textit{and, the, a, to, of,...} \\
         $\Tinsword{}$ &  \textbf{duplicate} a word \\
        $\Tsubword{}$ & \textbf{substitute} a word with one of its synonyms  \\
        \bottomrule
    \end{tabular}
    \label{tab:perturbation}
    \end{small}

\end{table}
Following~\citet{ZhangAD20}, we create perturbation spaces by combining the transformations in \Tblref{tab:perturbation}, e.g., $\SDS$ removes up to two stop words and replaces up to two words with synonyms.
We also design a domain-specific perturbation space $\reviewperturb{}$ for movie reviews; e.g., one transformation in $\reviewperturb{}$ can duplicate question or exclamation marks because they usually appear repeatedly in movie reviews.
We provide the detailed definition and evaluation of $\reviewperturb{}$ in the appendix.

For Tree-LSTMs, we consider the tree transformations exemplified in~\Figref{fig: treetrans} and described in \Secref{ssec:ext}.

\paragraph{Evaluation metrics} 
(1)
\emph{Accuracy} (Acc.) is the vanilla test accuracy.
(2)
\emph{HotFlip accuracy} (HF Acc.) is the adversarial accuracy with respect to the HotFlip attack~\cite{ebrahimi2018hotflip}: for each point in the test set, we use HotFlip to generate the top-5 perturbed examples and test if the classifications are correct on all the 5 examples and the original example.
%
% \aws{certification accuracy? for consistent terminology} 
(3)
\emph{Certified accuracy} (\Verifiedshort{}) 
is the percentage of points in the test set
certified as $S$-robust (\Eqref{eq:robustness}) using \toolname{}.
% \loris{explain what this means}
(4)
\emph{Exhaustive accuracy} (\Exhaustiveshort{}) is the 
percentage of points in the test set that is $S$-robust (\Eqref{eq:robustness}).
HotFlip accuracy is an upper bound of exhaustive accuracy; certified accuracy is a lower bound of exhaustive accuracy.

% In RQ2, we also use random smoothing~\cite{DBLP:conf/acl/YeGL20} to certify the accuracy of the model. 

% \aws{We might want to present this one later}

% \textbf{RS accuracy} 
%  is the accuracy certified by random smoothing~\cite{DBLP:conf/acl/YeGL20}: a prediction on $(\bfx, y)$ is considered correct if and only if more than half of perturbed samples in $S(\bfx)$ are correctly classified with $1\%$ significance level.

\paragraph{Baselines}
For training certifiable models against arbitrary string transformations, we compare \toolname{} to (1) \emph{Normal training} that minimizes the cross entropy. (2) \emph{Data augmentation} that augments the dataset with random samples from the perturbation space. (3) \emph{HotFlip~\cite{ebrahimi2018hotflip}} adversarial training.
And (4) \emph{A3T~\cite{ZhangAD20}} that trains robust CNNs.

For training certifiable models against word substitution, we compare \toolname{} to (1) \emph{\citet{DBLP:conf/emnlp/JiaRGL19}} that trains \emph{certifiably} robust (Bi)LSTMs. We call this \jiaapproach{} in the rest of this section. (2) \emph{ASCC~\cite{dong2021towards}} that trains \emph{empirically} robust (Bi)LSTMs. And (3) \emph{\citet{DBLP:conf/emnlp/HuangSWDYGDK19}} that trains \emph{certifiably} robust CNNs (comparison results are in the appendix).

For certification, we compare \toolname{} to (1) \emph{POPQORN~\cite{POPQORN}}, the state-of-the-art approach for certifying LSTMs. (2) \emph{SAFER~\cite{DBLP:conf/acl/YeGL20}} that provides probabilistic certificates to word substitution.

\emph{\citet{DBLP:conf/nips/XuS0WCHKLH20}} is a special case of \toolname{} where the perturbation space only contains substitution. We provide theoretical comparison in the appendix.

\begin{table}[t!]
    \centering
     \caption{Qualitative examples. The vanilla model incorrectly classifies the perturbed samples.}

    \renewcommand{\arraystretch}{1}
    \begin{small} 
    \begin{tabular}{p{6.2cm}c}
        \toprule
        \textit{Original sample in SST2 dataset}   \\
        i was perplexed to watch it unfold with an astonishing lack of passion or uniqueness {$\bm .$} & \textbf{-ve}\\
        \midrule
        \textit{Perturbed sample in $\SDS$}  \\
        i was perplexed \perturbed{\sout{to}} watch it unfold with an \perturbed{astounding}
        \perturbed{absence} of passion or uniqueness $\bm .$ & \textbf{+ve}\\
        \toprule
        \textit{Original sample in SST2 dataset} \\
        this is pretty dicey material $\bm .$ & \textbf{-ve}
        \\
        \midrule
        \textit{Perturbed sample in $\SIS$} \\
        this \perturbed{becomes} pretty \perturbed{pretty} dicey material $\bm .$ \perturbed{$\bm .$}  & \textbf{+ve}
      \\
        \bottomrule
    \end{tabular}
    \end{small}
    \label{tab:examples}

\end{table}

\subsection{Arbitrary Perturbation Spaces}

% We first compare \toolname{} to approaches that can handle arbitrary spaces: data augmentation, HotFlip, and A3T.
\begin{table*}[t]
    \centering

    % \aws{should we also highlight the \Verifiedshort{} Acc column, because that's our contribution? / also bold points for winners in all columns}
    % \aws{we can potentially combine the three tables into one mega table, as they all have the same structure} \yh{todo: distinguish LSTM, Tree-LSTM, and Bi-LSTM}
    \caption{Results of LSTM (on SST2), Tree-LSTM (on SST), and Bi-LSTM (on SST2) for three perturbation spaces. %We show the clean accuracy (Acc.), HotFlip accuracy (HF Acc.), certified accuracy (\Verifiedshort{}), and exhaustive accuracy (\Exhaustiveshort{}) of four different training methods. 
    }

    \setlength{\tabcolsep}{4.5pt}
    \renewcommand{\arraystretch}{0.8}
    \begin{small}
    \begin{tabular}{cl *3{cccc}}
        \toprule
         && \multicolumn{4}{>{\centering\arraybackslash}p{\lengththreecell{}}}{$\{(\Tdelword{}, 2),(\Tsubword{},2)\}$} & \multicolumn{4}{>{\centering\arraybackslash}p{\lengththreecell{}}}{$\{(\Tinsword{}, 2),(\Tsubword{},2)\}$} & \multicolumn{4}{>{\centering\arraybackslash}p{\lengththreecell{}}}{$\{(\Tdelword{}, 2),(\Tinsword{},2)\}$} \\ 
         
         \cmidrule(lr){3-6} \cmidrule(lr){7-10} \cmidrule(lr){11-14}
        &Train & {\scriptsize Acc.} & {\scriptsize HF Acc.} & {\scriptsize \Verifiedshort{}} & {\scriptsize \Exhaustiveshort{}} & {\scriptsize Acc.} & {\scriptsize HF Acc.} & {\scriptsize \Verifiedshort{}} & {\scriptsize \Exhaustiveshort{}} & {\scriptsize Acc.} & {\scriptsize HF Acc.} & {\scriptsize \Verifiedshort{}} & {\scriptsize \Exhaustiveshort{}} \\
        \midrule
        \multirow{4}{*}{\rotatebox[origin=c]{90}{\textsc{\tiny{LSTM}}}} & Normal & \textbf{84.6} & 71.9 & \cellcolor{Gray!10}~~4.6 & \cellcolor{Gray!10}68.9 & 84.6 & 64.0 & \cellcolor{Gray!10}~~0.5 & \cellcolor{Gray!10}55.1 & \textbf{84.6} & 73.7 & \cellcolor{Gray!10}~~1.2 & \cellcolor{Gray!10}65.2 \\
        
        & Data Aug. & 84.0 & 77.0 & \cellcolor{Gray!10}~~5.5 & \cellcolor{Gray!10}74.4 & \textbf{84.7} & 70.2 & \cellcolor{Gray!10}~~0.3 & \cellcolor{Gray!10}61.5 & 84.5 & 75.4 & \cellcolor{Gray!10}~~0.4 & \cellcolor{Gray!10}68.3 \\
        
        & HotFlip & 84.0 & \textbf{78.7} & \cellcolor{Gray!10}~~4.3 & \cellcolor{Gray!10}74.6 & 82.5 & \textbf{75.9} & \cellcolor{Gray!10}~~0.0 & \cellcolor{Gray!10}62.0 & 84.4 & \textbf{80.6} & \cellcolor{Gray!10}~~0.0 & \cellcolor{Gray!10}68.7 \\ 
        
        & \toolname{} & 82.5 & 77.8 & \cellcolor{Gray!10}\textbf{72.5} &  \cellcolor{Gray!10}\textbf{77.0} & 80.2 & 70.0 & \cellcolor{Gray!10}\textbf{55.4} & \cellcolor{Gray!10}\textbf{64.0} & 82.6 & 78.6 & \cellcolor{Gray!10}\textbf{69.4} &  \cellcolor{Gray!10}\textbf{74.6} \\

                \midrule
        \multirow{4}{*}{\rotatebox[origin=c]{90}{\textsc{\tiny{Tree-LSTM}}}} & Normal & \textbf{50.3} & 39.9 & \cellcolor{Gray!10}~~4.1 & \cellcolor{Gray!10}33.8 & \textbf{50.3} & 33.4 & \cellcolor{Gray!10}~~0.0 & \cellcolor{Gray!10}17.9 & \textbf{50.3} & 40.1 & \cellcolor{Gray!10}~~0.0 & \cellcolor{Gray!10}25.7\\
        
        & Data Aug. & 47.5 & 40.8 & \cellcolor{Gray!10}~~1.4 & \cellcolor{Gray!10}36.4 & 48.1 & 37.1 & \cellcolor{Gray!10}~~0.0 & \cellcolor{Gray!10}23.0 & 47.6 & 40.6 & \cellcolor{Gray!10}~~0.0 & \cellcolor{Gray!10}29.0\\
        
        & HotFlip & 49.5 & 43.4 & \cellcolor{Gray!10}~~1.6 & \cellcolor{Gray!10}38.4 & 48.7 & \textbf{39.5} & \cellcolor{Gray!10}~~0.0 & \cellcolor{Gray!10}29.0 & 49.5 & 42.7 & \cellcolor{Gray!10}~~0.0 & \cellcolor{Gray!10}32.1 \\ 
        
        & \toolname{} & 46.4 & \textbf{43.4} & \cellcolor{Gray!10}\textbf{30.9} & \cellcolor{Gray!10}\textbf{41.9} & 46.1 & 39.0 & \cellcolor{Gray!10}\textbf{17.1} & \cellcolor{Gray!10}\textbf{37.6} & 46.5 & \textbf{43.8} & \cellcolor{Gray!10}\textbf{19.2} & \cellcolor{Gray!10}\textbf{40.0}\\

                \midrule
        \multirow{4}{*}{\rotatebox[origin=c]{90}{\textsc{\tiny{Bi-LSTM}}}} & Normal & 83.0 & 71.1 & \cellcolor{Gray!10}~~8.2 & \cellcolor{Gray!10}68.0 & 83.0 & 63.4 & \cellcolor{Gray!10}~~2.1 &  \cellcolor{Gray!10}56.1 & 83.0 & 72.5 & \cellcolor{Gray!10}~~6.4 & \cellcolor{Gray!10}65.5 \\
        
        & Data Aug. & 83.2 & 75.1 & \cellcolor{Gray!10}~~8.7 & \cellcolor{Gray!10}72.9 & \textbf{83.5} & 66.8 & \cellcolor{Gray!10}~~1.3 & \cellcolor{Gray!10}59.1 & \textbf{84.6} & 75.0 & \cellcolor{Gray!10}~~4.6 & \cellcolor{Gray!10}68.6 \\
        
        & HotFlip & \textbf{83.6} & \textbf{79.2} & \cellcolor{Gray!10}~~9.2 & \cellcolor{Gray!10}73.4 & 82.8 & \textbf{76.6} & \cellcolor{Gray!10}~~0.1 & \cellcolor{Gray!10}55.5 & 83.5 & \textbf{79.1} & \cellcolor{Gray!10}~~0.0 & \cellcolor{Gray!10}55.7\\ 
        
        & \toolname{} & 83.5 & 78.7 & \cellcolor{Gray!10}\textbf{70.9} & \cellcolor{Gray!10}\textbf{77.5} & 80.2 & 71.4 & \cellcolor{Gray!10}\textbf{59.8} & \cellcolor{Gray!10}\textbf{66.4} & 82.6 & 76.2 & \cellcolor{Gray!10}\textbf{66.2} & \cellcolor{Gray!10}\textbf{71.8}\\ 
        \bottomrule
    \end{tabular}
    \end{small}
    \label{tab:lstm_whole}

\end{table*}

\begin{table}[t]
    \centering
    \caption{Results of LSTM on SST2 dataset for $\reviewperturb{}$. }
    \renewcommand{\arraystretch}{0.8}
    \begin{small}
    \begin{tabular}{l *1{p{\lengthcellshorter{}}p{\lengthcell{}}p{\lengthcell{}}p{\lengthcelllonger{}}}}
        \toprule
         & \multicolumn{4}{>{\centering\arraybackslash}p{\lengththreecell{}}}{$\reviewperturb{}$}\\ 
         
         \cmidrule(lr){2-5}
        Train  & {\small Acc.} & {\scriptsize HF Acc.} & {\scriptsize \Verifiedshort{}} & {\scriptsize \Exhaustiveshort{}}  \\
        \midrule
        Normal & \textbf{83.9} & 72.4 & 21.0 & 71.0 \\
        Data Aug. & 79.2 & 72.5 & 30.5 & 71.7\\
        HotFlip & 79.6 & 74.3 & 34.5 & 72.9\\
        \textbf{\toolname{}}  & 82.3 & \textbf{78.1} & \textbf{74.2} & \textbf{77.1}\\
        \bottomrule
    \end{tabular}
    \end{small}
    \label{tab:compare_DSreview}

\end{table}

\textbf{Comparison to Data Augmentation \& HotFlip.}
% We compare to these two baselines on four perturbation spaces
% shown in \twoTblrefs{tab:lstm}{tab:compare_DSreview}.
%$\{(\Tdelword{}, 2), (\Tsubword{}, 2)\}$, $\{(\Tinsword{},2), (\Tsubword{},2)\}$, $\{(\Tdelword{},2), (\Tinsword{},2)\}$, and $\reviewperturb{}$.
We use the three perturbation spaces in \Tblref{tab:lstm_whole} and the domain-specific perturbation space $\reviewperturb{}$ in \Tblref{tab:compare_DSreview}.

\textbf{\textit{\toolname{} 
 outperforms 
data augmentation and HotFlip in terms of \Exhaustiveshort{} and \Verifiedshort{}
}}
\Tblref{tab:lstm_whole} shows the results of LSTM, Tree-LSTM, and Bi-LSTM models on the tree perturbation spaces.
\Tblref{tab:compare_DSreview} shows the results of LSTM models on the domain-specific perturbation space $\reviewperturb{}$.
\toolname{} has significantly higher \Exhaustiveshort{}
than normal training ($+8.1, +14.0, +8.7$ on average), data augmentation ($+4.2, +10.4, +5.0$), and HotFlip ($+3.6, +6.7, +10.4$) for LSTM, Tree-LSTM, and Bi-LSTM respectively. 

Models trained with \toolname{} have a relatively high \Verifiedshort{} ($53.6$ on average). 
Data augmentation and HotFlip 
result in models not amenable to certification---in some cases, almost nothing in the test set can be certified.

\toolname{} produces more robust models at the expense of accuracy. 
Other robust training approaches like \jiaapproach{} and A3T also exhibit this trade-off.
However, as we will show next, \toolname{} retains higher accuracy than these approaches.
% Moreover, \toolname{} retains higher accuracy than data augmentation and HotFlip for $\reviewperturb{}$.

% \begin{table}[t]
%     \centering
%     \caption{\toolname{} vs A3T (CNN) on SST2 dataset. }
%     % \setlength{\tabcolsep}{1pt}
%     \vskip 0.1in
%     \small
%     \begin{tabular}{ll *1{p{\lengthcellshorter{}}p{\lengthcell{}}p{\lengthcell{}}p{\lengthcelllonger{}}}}
%         \toprule
%          && \multicolumn{4}{>{\centering\arraybackslash}p{\lengththreecell{}}}{$\{(\Tinsword{}, 2),(\Tsubword{},2)\}$} \\ 
         
%          \cmidrule(lr){3-6} 
%         Training & Model & {\small Acc.} & {\scriptsize HF Acc.} & {\scriptsize \Verifiedshort{}} & {\scriptsize \Exhaustiveshort{}} \\
%         \midrule
%         A3T & CNN & 79.9 & 68.3 & N/A & 57.7 \\
%         \toolname{} & LSTM & \textbf{80.2} & \textbf{70.0} & \textbf{55.4} & \textbf{64.0}\\
%         \bottomrule
%     \end{tabular}
%     \label{tab:compare_a3twhole}
% \end{table}
\begin{table*}[t]
    \centering
    \caption{\toolname{} vs A3T (CNN) on SST2 dataset. }

    \begin{small}
    \begin{tabular}{ll *2{p{\lengthcellshorter{}}p{\lengthcell{}}p{\lengthcell{}}p{\lengthcelllonger{}}}}
        \toprule
         && \multicolumn{4}{>{\centering\arraybackslash}p{\lengththreecell{}}}{$\{(\Tdelword{}, 2),(\Tsubword{},2)\}$} & \multicolumn{4}{>{\centering\arraybackslash}p{\lengththreecell{}}}{$\{(\Tinsword{}, 2),(\Tsubword{},2)\}$} \\ 
         
         \cmidrule(lr){3-6} \cmidrule(lr){7-10} 
        Train & Model & {\small Acc.} & {\scriptsize HF Acc.} & {\scriptsize \Verifiedshort{}} & {\scriptsize \Exhaustiveshort{}} & {\small Acc.} & {\scriptsize HF Acc.} & {\scriptsize \Verifiedshort{}} & {\scriptsize \Exhaustiveshort{}}  \\
        \midrule
        A3T (HotFlip) & CNN & 80.2 & 71.9 & N/A & 70.2 & 79.9 & 68.3 & N/A & 57.7 \\
        \toolname{} & LSTM & \textbf{82.5} & \textbf{77.8} & \textbf{72.5} & \textbf{77.0} & \textbf{80.2} & \textbf{70.0} & \textbf{55.4} & \textbf{64.0}\\
        \bottomrule
    \end{tabular}
    \end{small}
    \label{tab:compare_a3twhole}

\end{table*} 
\paragraph{Comparison to A3T}
% We compare \toolname{} against A3T using $\SDS$ and $\SIS$. 
% We do not use $\SDI$ because A3T degenerates to HotFlip training for this perturbation space.
A3T degenerates to HotFlip training on $\SDI$, so we do not use this perturbation space.
%We download two models from the A3T website for comparison.\footnote{\url{github.com/ForeverZyh/A3T}}

\textbf{\textit{
The LSTMs trained using \toolname{} are more robust than the CNNs trained by A3T for both perturbation spaces; \toolname{} can certify the robustness of models while A3T cannot.}}
\Tblref{tab:compare_a3twhole} shows that
\toolname{} results in models with higher accuracy ($+2.3$ and $+0.3$), HF Acc. ($+5.9$ and $+1.7$), and \Exhaustiveshort{} ($+6.8$ and $+6.3$) than those produced by A3T. 
\toolname{} can certify the trained models while A3T cannot.
% \Tblref{tab:compare_a3t} shows the results of the perturbation space $\SDS$ (results of $\SIS$ are in the appendix).

% \yh{move it here.} We can apply the idea of A3T to \toolname{}, extending \toolname{} to abstract any subset of the given perturbation space and to augment the remaining perturbation space. We show the effectiveness of this extension in the appendix.

\subsection{Experiments on Word Substitution}
We compare to
works \emph{limited} to word substitution.

% \loris{following para can still be cleaned}
\begin{table}[t]
    \centering
    \caption{ \toolname{} vs \jiaapproach{} and ASCC on IMDB dataset.}
    \renewcommand{\arraystretch}{0.8}
    \setlength{\tabcolsep}{5pt}

    \begin{small}
    \begin{tabular}{lllllll}
        \toprule
         & \multicolumn{3}{c}{$\{(\Tsubword{},1)\}$} & \multicolumn{3}{c}{$\{(\Tsubword{},2)\}$} \\ 
         
         \cmidrule(lr){2-4} \cmidrule(lr){5-7}
        {\scriptsize Train} & {\scriptsize Acc.} & {\scriptsize \Verifiedshort{}} & {\scriptsize \Exhaustiveshort{}} & {\scriptsize Acc.} & {\scriptsize \Verifiedshort{}} & {\scriptsize \Exhaustiveshort{}}  \\
        \midrule
        {\scriptsize \jiaapproach{}} & 76.8 & 67.0 & 71.0 & 76.8 & 64.8 &  68.3\\
        {\scriptsize ACSS} & 82.8 & ~~0.0 & 81.5 & 82.8 & ~~0.0 & \textbf{80.8}\\
        {\scriptsize \textbf{\toolname{}}} & \textbf{87.7} & \textbf{77.8} & \textbf{82.6} & \textbf{86.3} & \textbf{71.0} & 78.2\\
        \bottomrule
    \end{tabular}
    \end{small}
    \label{tab:compare_jia}

\end{table}

\paragraph{Comparison to \jiaapproach{} and ASCC} We choose two perturbation spaces, $\{(\Tsubword{}, 1)\}$ and $\{(\Tsubword{}, 2)\}$.
% \loris{following sentence is so confusing. Is it 2 models, 2 models per perturbations?}
We train one model per perturbation space using \toolname{} under the same experimental setup of \jiaapproach{}, BiLSTM on the IMDB dataset.
%provided in their paper, i.e., the same dataset (IMDB), synonym sets, and the Bi-LSTM architecture.
By definition, \jiaapproach{} and ASCC
train for an arbitrary number of substitutions.
\Verifiedshort{} is computed using \toolname{}. 
Note that \jiaapproach{} can only certify for $\{(\Tsubword{}, \infty)\}$ and ASCC cannot certify.
%We download the trained model from their website and evaluate it on $\{(\Tsubword{}, 1)\}$ and $\{(\Tsubword{}, 2)\}$.\footnote{\url{github.com/robinjia/certified-word-sub}}

\textbf{\textit{\toolname{} trains more robust models than \jiaapproach{} for two perturbation spaces with word substitution.}}
\Tblref{tab:compare_jia} shows that \toolname{} achieves higher accuracy, 
% ($+10.9$ and $+9.5$, respectively)
\Verifiedshort{}, 
% ($+10.8$ and $+6.2$, respectively)
and \Exhaustiveshort{} 
% ($+11.6$ and $+9.9$, respectively) 
than \jiaapproach{} on the two perturbation spaces.

% \paragraph{Comparison to ASCC.} We use the same perturbation spaces and the same model structure as in the comparison to \jiaapproach{}. 
%We train the model using their code.\footnote{\url{https://github.com/dongxinshuai/ASCC}}. 

\textbf{\textit{\toolname{} trains a more robust model than ASCC for $\{(\Tsubword{}, 1)\}$, but ASCC's model is more robust for $\{(\Tsubword{}, 2)\}$.}}
% \yh{results are not ready.}
% \toolname{} trains a more robust model ($+1.1$ \Exhaustiveshort{}) than \toolname{} for $\{(\Tsubword{}, 1)\}$, but ASCC's model is more robust (\yh{$+X.X$} \Exhaustiveshort{}) for $\{(\Tsubword{}, 2)\}$.
\Tblref{tab:compare_jia} shows that the \toolname{}-trained models have higher accuracy
%($+4.9$ and $+3.5$, respectively) 
and \Verifiedshort{}
%($+77.8$ and $+71.0$, respectively).

\paragraph{Comparison to POPQORN} We compare the certification of an \toolname{}-trained model and a normal model against $\{(\Tsubword{}, 3)\}$ on the first 100 examples in SST2 dataset. Because POPQORN can only certify the $l_p$ norm ball, we overapproximate the radius of the ball as the maximum $l_1$ distance between the original word and its synonyms. 

\textbf{\emph{\toolname{} runs much faster than POPQORN. \toolname{} is more accurate than POPQORN on the \toolname{}-trained model, while POPQORN is more accurate on the normal model.}} \toolname{} certification takes 0.17sec/example on average for both models, while POPQORN certification takes 12.7min/example. \toolname{} achieves 67\% and 5\% \Verifiedshort{} on \toolname{}-trained model and normal model, respectively, while POPQORN achieves 22\% and 28\% \Verifiedshort{}, but crashes on 45\% and 1\% of examples for two models.

\paragraph{Comparison to SAFER~\cite{DBLP:conf/acl/YeGL20}} 
SAFER is a post-processing technique for
certifying robustness via \emph{randomized smoothing}.
% Therefore, we can apply it to \toolname{} models.
% We compute the randomized smoothing accuracy using the perturbation spaces $\{(\Tsubword{}, 1)\}$ and $\{(\Tsubword{}, 2)\}$---randomized smoothing has not been studied for perturbation spaces involving deletions and insertions.
%
%
We train a Bi-LSTM model using \toolname{}
following SAFER's experimental setup on the IMDB dataset
and SAFER's synonym set, which is different from \jiaapproach{}'s.
We consider the perturbation spaces $\{(\Tsubword{}, 1)\}$ and $\{(\Tsubword{}, 2)\}$.
We use both \toolname{} and SAFER to certify the robustness.
The significance level of SAFER is set to $1\%$.

\textbf{\emph{SAFER has a higher certified accuracy than \toolname{}. However, 
its certificates are statistical, tied to word substitution only,
and are slower to compute.}}
Considering $\{(\Tsubword{}, 2)\}$, \toolname{} results in a certified accuracy of 79.6 and SAFER results in a certified accuracy of 86.7 (see appendix). 
Note that certified accuracies are incomparable because SAFER computes certificates
that only provide statistical guarantees.
% However, we observe higher randomized smoothing accuracy
% by applying SAFER to models trained using data augmentation instead of ARC (e.g., 89.7 vs. 86.7).
Also, note that ARC uses $O(n\prod_{i=1}^n \delta_i)$ forward passes for each sample, while SAFER needs to randomly sample thousands of times.  
In the future, it would be interesting to explore 
extensions of SAFER to \toolname{}'s rich perturbation spaces.

\subsection{ Effects of Perturbation Space Size} 
\begin{figure}[t]
    \centering
    % \vspace{-4mm}
    % \aws{y axis: "size of perturbation space"}
     \includegraphics[width=0.75\columnwidth]{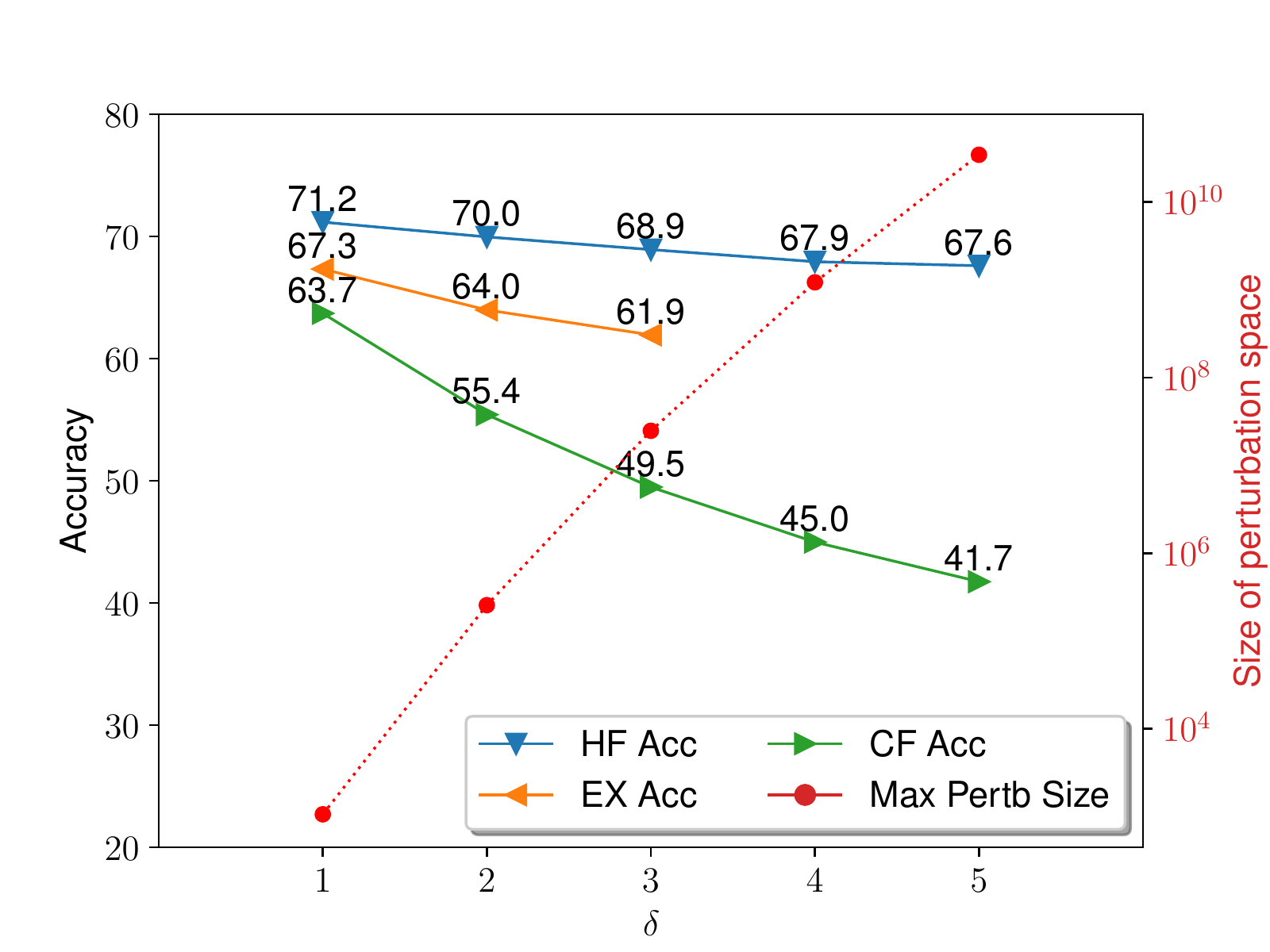}
    \caption{Accuracy metrics and the maximum perturbation space size as $\delta$ varies from 1 to 5 for $\{(\Tinsword{},\delta), (\Tsubword{},\delta)\}$.}
    \label{fig:RQ3}
\end{figure}

We fix the LSTM model (\Tblref{tab:lstm_whole}) trained using \toolname{} on $\SIS$ and SST2.
Then, we test this model’s \Verifiedshort{} and HotFlip accuracy on $\{(\Tinsword{},\delta), (\Tsubword{},\delta)\}$ by varying $\delta$ from 1 to 5.
We evaluate the \Exhaustiveshort{} only for $\delta=1,2,3$, due to the combinatorial  explosion. 

\textbf{\textit{\toolname{} maintains a reasonable \Verifiedshort{} for increasingly larger spaces}.} \Figref{fig:RQ3} shows the results, along with the maximum perturbation space size in the test set.
\toolname{} can certify $41.7\%$ of the test set even when the perturbation space size grows to about $10^{10}$.
For $\delta=1,2,3$, the \Verifiedshort{} is lower than the \Exhaustiveshort{} ($8.2$ on average), while the HF Acc. is higher than the \Exhaustiveshort{} ($5.6$ on average).
Note that \toolname{} uses a small
amount of time to certify the entire test set, $3.6$min on average,
using a single V100 GPU,
making it incredibly efficient compared
to brute-force enumeration.
% \aws{check}
% than HotFlip attack ($14.4$ minutes on average) to compute the bound. 
% \aws{give the amount of time it takes ARC to do verification}
\section{Conclusion}
We present \toolname{}, which uses memoization and abstract interpretation
to certify robustness to programmable perturbations for LSTMs. 
\toolname{} can be used to  train models that are more robust 
than those trained  using existing techniques and handle more complex perturbation spaces.
Last, the models trained with \toolname{} have high certification accuracy,
which can be certified using \toolname{} itself.
% and (3) \toolname{} can certify robustness to combinations of string transformations that are out-of-scope for existing techniques.

\section*{Acknowledgements}
We thank Nick Giannarakis and the anonymous reviewers for commenting on earlier drafts.
This work is supported by the
National Science Foundation grants CCF-1420866, CCF-1704117, CCF-1750965, CCF-1763871, CCF-1918211, CCF-1652140, the Microsoft Faculty Fellowship, and gifts and awards from Facebook.
\clearpage

\bibliography{main.bib}
\bibliographystyle{acl_natbib}

\clearpage
\appendix

% \input{sections/bridge.tex}

% \section{Appendix}
% \input{sections/appendix.tex}

% \textbf{\emph{Do not put content after the references.}}
% %
% Put anything that you might normally include after the references in a separate
% supplementary file.

% We recommend that you build supplementary material in a separate document.
% If you must create one PDF and cut it up, please be careful to use a tool that
% doesn't alter the margins, and that doesn't aggressively rewrite the PDF file.
% pdftk usually works fine. 

% \textbf{Please do not use Apple's preview to cut off supplementary material.} In
% previous years it has altered margins, and created headaches at the camera-ready
% stage. 
%%%%%%%%%%%%%%%%%%%%%%%%%%%%%%%%%%%%%%%%%%%%%%%%%%%%%%%%%%%%%%%%%%%%%%%%%%%%%%%
%%%%%%%%%%%%%%%%%%%%%%%%%%%%%%%%%%%%%%%%%%%%%%%%%%%%%%%%%%%%%%%%%%%%%%%%%%%%%%%

\appendix
\section{Appendix}

\subsection{Proof of Lemmas and Theorems}

\noindent \textbf{\Lemmaref{lemma:4.2}}
\begin{proof}
We prove \Lemmaref{lemma:4.2} by induction on $i,j$ and $S$. 

\textbf{Base case:} $H_{0,0}^\varnothing = {h_0}$ is defined by both \Eqref{eq: complexstate} and \Eqref{eq:contransition1}.

\textbf{Inductive step for $H_{i,j}^{S}$:} Suppose the lemma holds for $H_{i',j'}^{S'}$, where $(0\le i' \le i \wedge 0 \le j' \le j \wedge S' \subseteq S) \wedge (i' \neq i \vee j' \neq j \vee S' \neq S)$. 
$S' \subseteq S$ denotes that for all $(T_k, \delta_k') \in S'$, we have $(T_k, \delta_k) \in S$ and $\delta_k' \le \delta_k$. 
%And $S' \subset S$ denotes that $S' \subseteq S$ and there exists a transformation $T_k$ such that $(T_k, \delta_k') \in S' \wedge (T_k, \delta_k) \in S$ and $\delta_k' < \delta_k$.

$H_{i,j}^{S}$ in \Eqref{eq: complexstate} comes from two cases illustrated in \Figref{fig:illu_state_new}. These two cases are captured by the first line and the second line in \Eqref{eq:contransition1}, respectively. The inductive hypothesis shows that the lemma holds on states $H_{i-1,j-1}^{S}$ and $H_{i{-}t_k,j{-}s_k}^{S_{k\downarrow{}}}$. Thus, the lemma also holds on $H_{i,j}^{S}$.
\end{proof}

\noindent \textbf{\Theref{theorem: theorem4.1}}
\begin{proof}
We can expand \Eqref{eq: theF} using the decomposition of the perturbation space as
\vskip -0.1in
\begin{small}
\begin{align}
    &F \nonumber \\
    = &\bigcup_{S'\in \tightperturbset{S}} \{\lstm(\bfz, \hstate_0) \mid \bfz \in S'^{=}(\bfx)\}\nonumber \\
    =& \bigcup_{S'\in \tightperturbset{S}} \bigcup_{i \ge 0}\{\lstm(\bfz, \hstate_0)\mid \bfz \in S'^=(\bfx) \wedge \len{\bfz}=i\}
    \label{eq: theorem4.1_1}
\end{align}
\end{small}%

\Eqref{eq: theorem4.1_1} and \Eqref{eq:hfinal} are equivalent, leading to the equivalence of \Eqref{eq: theF} and \Eqref{eq:hfinal}.
\end{proof}

\noindent \textbf{\Theref{theorem: theorem4.2}}
\textbf{\Theref{theorem: theorem4.2}}

\begin{proof}
We first show that \Eqref{eq: finalstateoftight} is equivalent to 
\begin{align*}
   &\bigcup_{0 \le i \le \maxlen{\bfx}} H_{i,\len{\bfx}}^S \quad \text{, where }\\
   &\maxlen{\bfx} = \len{\bfx} + \sum_{(T_k, \delta_k) \in S} \max(t_k {-} s_k, 0)\delta_k
\end{align*}
where $f_k: \alphabet^{s_k} \to 2^{\alphabet^{t_k}}$. 
As we will prove later, $\maxlen{\bfx}$ is the upper bound of the length of the perturbed strings.
Because $t_k,s_k,\delta_k$ are typically small constants, we can regard $\maxlen{\bfx}$ as a term that is linear in the length of the original string $\len{\bfx}$, i.e., $\maxlen{\bfx} = O(\len{\bfx})$.

Now, we prove that $\maxlen{\bfx}$ is the upper bound of the length of the perturbed string. The upper bound $\maxlen{\bfx}$ can be achieved by applying all string transformations that increase the perturbed string's length and not applying any string transformations that decrease the perturbed string's length. Suppose a string transformation $T_k = (f_k, \varphi_k), f_k: \alphabet^{s_k} \to 2^{\alphabet^{t_k}} $ can be applied up to $\delta_k$ times, then we can apply it $\delta_k$ times to increase the perturbed string's length by $(t_k - s_k)\delta_k$.

The proof of soundness follows immediately from 
the fact that $\alpha$, $\sqcup$, and $\ibp{\lstm}$
overapproximate their inputs,
resulting in an overapproximation $F$.

The proof of complexity follows the property that the number of non-empty hyperrectangles $\ibp{H}_{i,j}^S$ is $O(\len{\bfx} \cdot \prod_{i=1}^n \delta_i)$. This property follows the definition of the string transformations and the tight perturbation space $\tightperturb{}$. $\ibp{H}_{i,j}^S$ can be non-empty iff 
\vskip -0.1in
\begin{small}
\[i = j + \sum_{(T_k, \delta_k) \in S} (t_k - s_k)\delta_k, \quad \text{ where } f_k: \alphabet^{s_k} \to 2^{\alphabet^{t_k}}\]
\end{small}%
For each $\ibp{H}_{i,j}^S$, we need to enumerate through all transformations, so the complexity is $O(\len{\bfx}\cdot n\prod_{i=1}^n \delta_i)$ in terms of the number of LSTM cell evaluations. The interval bound propagation needs only two forward passes for computing the lower bound and upper bound of the hyperrectangles, so it only contributes constant time to complexity. 
In all, the total number of LSTM cell evaluations needed is $O(\len{\bfx}\cdot n\prod_{i=1}^n \delta_i)$.
\end{proof}

\subsubsection{Comparison to \citet{DBLP:conf/nips/XuS0WCHKLH20}} 
The dynamic programming approach proposed in \citet{DBLP:conf/nips/XuS0WCHKLH20} is a special case of ARC where the perturbation space only contains substitutions. The abstract state $g_{i,j}$ in their paper (Page 6, Theorem 2) is equivalent to $\ibp{H}_{i,i}^{\{(\Tsubword{}, j)\}}$ in our paper. 

% \aws{rename section to something more descriptive, e.g., handling attention etc.}
\subsection{Handling Attention}
We have introduced the memoization and abstraction of final states in \Secref{sec:approach}.
Moreover, for LSTM architectures that compute attention of each state $h_i$, we would like to compute the interval abstraction of each state at the $i$th time step, denoted as $\ibp{H}_i$.

It is tempting to compute $H_i$ as 
\begin{align}
    H_i = \bigcup_{S'\in \tightperturbset{}} \bigcup_{0 \le j \le \len{\bfx}} H_{i,j}^{S'} \label{eq: tempted}
\end{align}
Unfortunately, \Eqref{eq: tempted} does not contain states that are in the middle of a string transformation (see next example).

\begin{example}
Consider the string transformation $T_\emph{swap}$ that swaps two adjacent words and suppose $\bfx = \text{``to the''}$, $S=\{(T_\emph{swap}, 1)\}$, then $H_1 =\{\lstm(\text{``to''}, \hstate_0), \lstm(\text{``the''}, \hstate_0)\}$.
However, the only non-empty state with $i=1$ is $H_{1,1}^\varnothing = \{\lstm(\text{``to''}, \hstate_0)\}$. The state $\lstm(\text{``the''}, \hstate_0)$ is missing because it is in the middle of transformation $T_\emph{swap}$.

But \Eqref{eq: tempted} is correct for $i=2$ because the transformation $T_\emph{swap}$ completes at time step 2.
\end{example}

Think of the set of all strings
in a perturbation space as a tree, like in \cref{fig:approachfig}(b),
where strings that share prefixes
share LSTM states.
We want to characterize a subset $G_{i,j}^S$ of LSTM
states at the $i$th layer where the perturbed prefixes have had \emph{all} transformations in a space $S$ applied on the original prefix $\substring{\bfx}{1}{j}$ and are in the middle of transformation $T_k$.
Intuitively, $G_{i,j}^S$ is a super set of $H_{i,j}^S$ defined in \Secref{sec:approach}.

We formally define  $G_{i,j}^S$ as follows:
\vskip -0.1in
\begin{small}
\begin{align}
    G_{i,j}^S = \{\lstm(\substring{\bfz}{1}{i}, \hstate_0)\mid \bfz \in \tightperturb{}(\substring{\bfx}{1}{j}) \wedge L_{\bfz}\ge i\} \label{eq: defineG1}
\end{align}
\end{small}%

We rewrite \Eqref{eq: defineG1} by explicitly applying the transformations defining the perturbation space $S$, thus deriving our final equations:
\vskip -0.1in
\begin{small}
\begin{align}
   G_{i,j}^S = & \bigcup_{1\leq k\leq n} \bigcup_{\substack{1\leq l \leq t_j\\ \varphi_k(\substring{\bfx}{a}{b})=1}} \left\{\! 
   \begin{aligned}
      \lstm(\textbf{c}, h) \mid & h \in G_{i-l, j-s_k}^{S_{k\downarrow{}}} \\
      & \textbf{c} \in f_{k,:l}(\substring{\bfx}{a}{b})
   \end{aligned}
   \right\} \nonumber \\
  &~~~~ \cup\ \{\lstm(x_j, h) \mid h \in G_{i-1,j-1}^S\} \label{eq: defineG2}
\end{align}
\end{small}%
% \vskip -0.1in
% \begin{small}
% \begin{align}
%   &\forall 1\le j \le n.  \nonumber\\
%   &~~~~G_{i,j,0}^S =\begin{cases}
%     \left\{\! 
%   \begin{aligned}
%       \lstm(c, h) \mid & h \in G_{i-1,0}^{S_{-j}} \\
%       & c \in f_{j,0}(\substring{\bfx}{a}{b})
%   \end{aligned}
%   \right\} &  \varphi_j(\substring{\bfx}{a}{b})=1\\
%   \varnothing & \text{otherwise}
%   \end{cases} \label{eq: defineG2_1} \\
%   \nonumber\\
%   &\forall 1\le j \le n \wedge 1\le k \le t_j.  \nonumber\\
%   &~~~~G_{i,j,k}^S =\begin{cases}
%     \left\{\! 
%   \begin{aligned}
%       \lstm(c, h) \mid & h \in G_{i-1,j,k-1}^S \\
%       & c \in f_{j,k}(\substring{\bfx}{a}{b})
%   \end{aligned}
%   \right\} &  \varphi_j(\substring{\bfx}{a}{b})=1\\
%   \varnothing & \text{otherwise}
%   \end{cases} \label{eq: defineG2_2} \\
%   \nonumber\\
%   & G_{i,0}^S = \bigcup_{\substack{1\leq j \leq n\\ \varphi_j(\substring{\bfx}{a}{b})=1}} \{\lstm(x_{i - \offset_S}, h) \mid h \in G_{i-1,j,t_j}^S\} \cup \nonumber \\
%   &~~~~~~~~~~~~\{\lstm(x_{i - \offset_S}, h) \mid h \in G_{i-1,0}^S\} \label{eq: defineG2_3} \\
%   & G_{i}^S = \bigcup_{\substack{1\leq j \leq n\\ 0\le k \le t_j}} G_{i,j,k}^S \cup G_{i,0}^S \label{eq: defineG2}\\
%   &H_i = \bigcup_{S' \in \tightperturbset{S}} G_i^{S'}
% \end{align}
% \end{small}%
where $a=j-s_k+1$ and $b=j$. Notation $f_{k,:l}(\substring{\bfx}{a}{b})$ collects the first $l$ symbols for each $\bfz$ in $f_{k}(\substring{\bfx}{a}{b})$, i.e.,
\[f_{k,:l}(\substring{\bfx}{a}{b}) =\{\substring{\bfz}{1}{l} \mid \bfz \in f_{k}(\substring{\bfx}{a}{b})\}\]

$G_{i,j}^S$ contains (1) strings whose suffix is perturbed by $T_k = (\varphi_k,f_k)$ and the last symbol of $\bfz$ is the $l$th symbol of the output of $T_k$ (the first line of \Eqref{eq: defineG2}), and (2) strings whose suffix (the last character) is not perturbed by any transformations (the second line of \Eqref{eq: defineG2}).

Then, $H_i$ can be defined as
\begin{align*}
    H_i = \bigcup_{S'\in \tightperturbset{}} \bigcup_{0\le j \le \len{\bfx}} G_{i,j}^{S'}
\end{align*}

\begin{lemma}
\Eqref{eq: defineG1} and \Eqref{eq: defineG2} are equivalent.
\end{lemma}
The above lemma can be proved similarly to \Lemmaref{lemma:4.2}. 
%\Eqref{eq: defineG2} over-approximates \Eqref{eq: defineG1} because $h, c$ may not match in the first line, e.g., $h$ is a state that is in the middle of transformation $T_{j_1}$, but $c$ is from $f_{j_2,k}$.
%This over-approximation can be fixed by restricting the definition of 

We use interval abstraction to abstract \Eqref{eq: defineG2} similarity as we did in \Secref{sec:abstract}. 
The total number of LSTM cell evaluation needed is $O(\len{\bfx}\cdot \max_{i=1}^n(t_i) \cdot n\prod_{i=1}^n \delta_i)$ .

\subsection{Handling Bi-LSTMs}
Formally, We denote $\bfx^\reverse$ as the reversed string $\bfx$.
Suppose a transformation $T$ has a match function $\varphi$ and a replace function $f$, the reversed transformation $T^\reverse = (\varphi^\reverse, f^\reverse)$ is defined as 
\vskip -0.1in
\begin{small}
\[\varphi^\reverse(\bfx) = \varphi(\bfx^\reverse), f^\reverse(\bfx) = \{\bfz^\reverse \mid \bfz \in f(\bfx^\reverse)\}\quad  \forall \bfx \in \alphabet^*\]
\end{small}%

\subsection{Handling Tree-LSTMs}
Intuitively, we replace substrings in the formalization of LSTM with subtrees in the Tree-LSTM case. 
We denote the subtree rooted at node $u$ as $\bft_u$ and the size of $\bft_u$ as $\size{\bft_u}$.
The state $H_u^S$ denotes the Tree-LSTM state that reads subtree $\bft_u$ generated by a tight perturbation space $S$. The initial states are the states at leaf node $u$
\[H_u^\varnothing = \{\lstm(x_u, \hstate_0)\} \]
and the final state is $H_\emph{root}^S$. 

%Instead of giving transition equations for general transformations like we did for LSTM and Bi-LSTM, 
We provide transition equations for three specific tree transformations \Figref{fig: treetrans}.

\subsubsection{Merge states}
For a non-leaf node $v$, we will merge two states, each from a child of $v$.
\vskip -0.1in
\begin{small}
\begin{align}
    H_v^S = \bigcup_{S' \in \tightperturbset{S}}\{\treelstm(h, g)\mid h\in H_{v_1}^{S'} \wedge g\in H_{v_2}^{S-S'}\} \label{eq: mergetree}
\end{align}
\end{small}%
where $v_1$ and $v_2$ are children of $v$, and $\treelstm$ denotes the Tree-LSTM cell that takes two states as inputs.
Notation $S - S'$ computes a tight perturbation space by subtracting $S'$ from $S$. Formally, suppose 
\begin{align*}
    S &=\{(T_1, \delta_1), (T_2, \delta_2),\ldots, (T_n, \delta_n)\}\\
    S'&=\{(T_1, \delta'_1), (T_2, \delta'_2),\ldots, (T_n, \delta'_n)\}
\end{align*}
then 
\vskip -0.1in
\begin{small}
\[S-S' = \{(T_1, \delta_1-\delta'_1), (T_2, \delta_2-\delta'_1),\ldots, (T_n, \delta_n - \delta'_n)\}\]
\end{small}%
Notice that \Eqref{eq: mergetree} is general to any tight perturbation space $S$ containing these three tree transformations.
% We use \Eqref{eq: mergetree} to merge states at non-leaf nodes for any tight perturbation space.
\begin{table*}[t]
    \centering
     \caption{String transformations for $\reviewperturb{}$.}

    % \setlength{\tabcolsep}{1pt}
    % \rowcolors{2}{}{gray!10}
    \begin{small}
    \begin{tabular}{ll}
        \toprule
         Trans & Description\\
        \midrule
          $\TReviewFirst{}$ & substitute a phrase in the set $A$ with another phrase in $A$.\\
         $\TReviewSecond{}$ & substitute a phrase in the set $B$ with another phrase in $B$ or substitute a phrase in $C$ with another phrase in $C$.\\
        $\TReviewThird{}$ & delete a phrase ``one of'' from ``one of the most ...'' or from ``one of the ...est''.  \\
        $\TReviewFourth{}$ & duplicate a question mark ``?'' or an exclamation mark ``!''.\\
        \bottomrule
    \end{tabular}
    \label{tab:s_review}
    \end{small}

\end{table*}

\subsubsection{$\Tsubword{}$}
We first show the computation of $H_u^S$ for a leaf node $u$. The substitution only happens in the leaf nodes because only the leaf nodes correspond to words.
\[
    H_u^{\{\Tsubword{}, 1\}} = \{\lstm(c, \hstate_0)\mid c\in f_\emph{sub}(x_u)\}
\]

\subsubsection{$\Tinsword{}$}
$\Tinsword{}$ can be seen as a subtree substitution at leaf node $u$.
\[
    H_u^{\{\Tinsword{}, 1\}} = \{\treelstm(h, h)\},
\]
where $h=\lstm(x_u, h_0)$.

\subsubsection{$\Tdelword{}$}
Things get tricky for $\Tdelword{}$ because $\{(\Tdelword{}, \delta)\}$ can delete a whole subtree $\bft_v$ if (1) the subtree only contains stop words and (2) $\size{\bft_v} \le \delta$. 
We call such subtree $\bft_u$ \textit{deletable} if both (1) and (2) are true.

Besides the merging equation \Eqref{eq: mergetree}, we provide another transition equation for $H_v^S$, where $v$ is any non-leaf node with two children $v_1, v_2$ and a perturbation space $S = \{\ldots, (\Tdelword{}, \delta) ,\ldots\}$
\vskip -0.1in
\begin{small}
\begin{align}
    H_v^S = &\bigcup_{S' \in \tightperturbset{S}}\{\treelstm(h, g)\mid h\in H_{v_1}^{S'} \wedge g\in H_{v_2}^{S-S'}\} \nonumber\\
    ~~~~~~~~&\cup \begin{cases}
    H_{v_2}^{S - S_\emph{del}^{(1)}} & \text{(1) $\bft_{v_1}$ is \textit{deletable}}\\
    H_{v_1}^{S - S_\emph{del}^{(2)}} & \text{(2) $\bft_{v_2}$ is \textit{deletable}}\\
    H_{v_2}^{S - S_\emph{del}^{(1)}} \cup H_{v_1}^{S - S_\emph{del}^{(2)}} & \text{both (1) and (2)}\\
    \varnothing & \text{otherwise}
    \end{cases}
\end{align}
\end{small}%
where 
\vskip -0.1in
\begin{small}
\[S_\emph{del}^{(1)} = \{(\Tdelword{}, \size{\bft_{v_1}})\}, S_\emph{del}^{(2)} = \{(\Tdelword{}, \size{\bft_{v_2}})\}\]
\end{small}%

\subsubsection{Soundness and Complexity}

We use interval abstraction to abstract the transition equations for Tree-LSTM similarity as we did in \Secref{sec:abstract}. 
The total number of LSTM/Tree-LSTM cell evaluations needed is $O(\size{\bft}(\prod_{i=1}^n \delta_i)^2)$.
The term $(\prod_{i=1}^n \delta_i)^2$ comes from \Eqref{eq: mergetree}, as we need to enumerate $S'$ for each $S$ in the decomposition set.

\subsection{Experimental Setup}

We conduct all experiments on a server running Ubuntu 18.04.5 LTS with V100 32GB GPUs and Intel Xeon Gold 5115 CPUs running at 2.40GHz.

\subsubsection{Definition for $\reviewperturb{}$}
We design $\reviewperturb{}$ by inspecting highly frequent n-grams in the movie review training set. Formally, 
\begin{align*}
    \reviewperturb{} = \{&(\TReviewFirst{}, 1), (\TReviewSecond{}, 1), \\
    &(\TReviewThird{}, 1), (\TReviewFourth{}, 1), (\Tsubword{}, 2)\}
\end{align*}
where $\TReviewFirst{}$, $\TReviewSecond{}$, $\TReviewThird{}$, and $\TReviewFourth{}$ are defined in \Tblref{tab:s_review} with
\begin{align*}
    A = \{&\text{``this is''}, \text{``this 's''}, \text{``it is''}, \text{``it 's''}\}\\
    B = \{&\text{``the movie''}, \text{``the film''}, \text{``this movie''}, \\
    &\text{``this film''}, \text{``a movie''}, \text{``a film''}\}\\
    C = \{&\text{``the movies''}, \text{``the films''}, \text{``these movies''}, \\
    & \text{``these films''}\}
\end{align*}

\subsubsection{Implementation of \toolname{}} 
We provide a general implementation of \toolname{} on LSTM against arbitrary user-defined string transformations.
We also provide specific implementations of \toolname{} on LSTM, Tree-LSTM, and Bi-LSTM against three transformations in \Tblref{tab:perturbation} and \Figref{fig: treetrans}.
Specific transformations allow us to optimize the specific implementations to utilize the full power of parallelism on GPU so that the specific implementations are faster than the general implementation. 
We conduct all our experiments on the specific implementations except for $\reviewperturb{}$.

% Instead of \Eqref{eq: defineG2}, the general implementation of LSTM computes $G_{i,j}^S$ as 
% \vskip -0.1in
% \begin{small}
% \begin{align*}
%   G_{i,j}^S = & \bigcup_{1\leq k\leq n} \bigcup_{\substack{1\leq l \leq t_j\\ \varphi_k(\substring{\bfx}{a}{b})=1}} \left\{\! 
%   \begin{aligned}
%       \lstm(\textbf{c}, h) \mid & h \in G_{i-l, j-s_k}^{S_{k\downarrow{}}} \\
%       & \textbf{c} \in f_{k,l}(\substring{\bfx}{a}{b})
%   \end{aligned}
%   \right\} \\
%   &~~~~ \cup\ \{\lstm(x_j, h) \mid h \in G_{i-1,j-1}^S\} 
% \end{align*}
% \end{small}%
% Notation $f_{j,k}(\substring{\bfx}{a}{b})$ collects the $k$th symbol for each $\bfz$ in $f_{j}(\substring{\bfx}{a}{b})$, i.e.,
% \[f_{j,k}(\substring{\bfx}{a}{b}) =\{z_k \mid \bfz \in f_{j}(\substring{\bfx}{a}{b})\}\]
% This implementation leads to the overapproximation of \Eqref{eq: defineG1}, and we leave the implementation of \Eqref{eq: defineG2} to future work.

\subsubsection{Details of Training}
\textit{A3T:} A3T has two instantiations, A3T (HotFlip) and A3T (Enum).
The difference between the two instantiations is the way it explores the augmentation space in A3T.
We choose to show A3T (HotFlip) for comparison, but \toolname{} wins over A3T (Enum) as well.

\textit{ASCC:} ASCC updates the word embedding during training by defaults. In our experiments, we fix the word embedding for ASCC.

\textit{\toolname{}:} All the models trained by \toolname{} have hidden state and cell state dimensions set to 100. 
We adopt a curriculum-based training method~\cite{DBLP:conf/iccv/GowalDSBQUAMK19, DBLP:conf/emnlp/HuangSWDYGDK19, DBLP:conf/emnlp/JiaRGL19, ZhangAD20} for training \toolname{} by using a hyperparameter $\lambda$ to weigh between the normal loss and the abstract loss and using a hyperparameter $\epsilon$ to gradually increasing the radius of synonym sets.
We gradually change two hyperparameters from 0 to their maximum values by $T_1$ epochs and keep training with their maximum values by $T_2$ epochs.

\textbf{Maximum values of hyperparameters $\lambda$ and $\epsilon$.}
For the experiments in \Tblref{tab:lstm_whole}, we tune the maximum of $\lambda$ during training from $0.5$ to $1.0$ (with span $0.1$) for LSTM and Bi-LSTM models and from $0.05$ to $0.10$ (with span $0.01$) for Tree-LSTM models.
For other experiments, which only use word substitutions, we fix the maximum of $\lambda$ to be $0.8$ following~\citet{DBLP:conf/emnlp/JiaRGL19}.

For every experiment, the maximum of $\epsilon$ during training is defined by the size of word substitutions in the perturbation space. For example, $\{(\Tdelword{}, 2),(\Tsubword{},2)\}$ defines the maximum of $\epsilon$ as $2$ and $\{(\Tdelword{}, 2),(\Tinsword{},2)\}$ defines the maximum of $\epsilon$ as $0$.

\textbf{Epoch numbers $T_1$ and $T_2$.} For LSTM and Bi-LSTM models on SST2 dataset, we set $T_1=16, T_2=8$. 
For other models (Tree-LSTM models on SST dataset and Bi-LSTM models on IMDB dataset), we set $T_1=20, T_2=10$. 

We use early stopping for other training methods and step the early stopping epoch as 5.

We provide the training scripts and all trained models in supplementary materials.

\begin{table}[t]
    \centering
    \caption{\toolname{} vs SAFER  on IMDB dataset.}

    \begin{small}
    \begin{tabular}{lllll}
        \toprule
         & \multicolumn{2}{c}{$\{(\Tsubword{},1)\}$} & \multicolumn{2}{c}{$\{(\Tsubword{},2)\}$} \\ 
         
         \cmidrule(lr){2-3} \cmidrule(lr){4-5}
        Train & \Verifiedshort{} & RS Acc. & \Verifiedshort{} & RS Acc. \\
        \midrule
        % Smoothing & \textbf{89.8} & ~~0.2 & 82.9 & \textbf{90.0} & \textbf{89.9} & ~~0.1 & 79.9$^*$ & \textbf{89.7}\\
        % \toolname{} & 86.8 & \textbf{82.0} & \textbf{84.8} & 87.2 & 86.4 & 
        % \textbf{79.6} & \textbf{83.0} & 86.7\\
        Data Aug. & ~~0.2 & \textbf{90.0} & ~~0.1 & \textbf{89.7}\\
        \toolname{} & \textbf{82.0} & 87.2 & 
        \textbf{79.6} & 86.7\\
        \bottomrule
    \end{tabular}
    \end{small}
    \label{tab:compare_rswhole}

\end{table}

\subsection{Evaluation Results}
The full results of comparison to SAFER are shown in \Tblref{tab:compare_rswhole}.

\paragraph{Comparison to \citet{DBLP:conf/emnlp/HuangSWDYGDK19}} We use $\{(\Tsubword{}, 3)\}$ on SST2 dataset for comparison between \toolname{} and \citet{DBLP:conf/emnlp/HuangSWDYGDK19}. We directly quote the results in their paper.

\begin{table}[t]
    \centering
    \caption{\toolname{} vs \citet{DBLP:conf/emnlp/HuangSWDYGDK19} (CNN) on SST2 dataset. }
    \setlength{\tabcolsep}{5pt}

    \begin{small}
    \begin{tabular}{ll *1{cccc}}
        \toprule
         && \multicolumn{4}{c}{$\{(\Tsubword{},3)\}$}\\ 
         
         \cmidrule(lr){3-6}
        Train & Model & {\small Acc.} & {\scriptsize HF Acc.} & {\scriptsize \Verifiedshort{}} & {\scriptsize \Exhaustiveshort{}}  \\
        \midrule
        Huang et al. & CNN & 81.7 & 77.2 & 44.5 & 76.5 \\
        \toolname{} & LSTM & \textbf{83.3} & \textbf{78.3} & \textbf{73.3} & \textbf{77.9}\\
        \bottomrule
    \end{tabular}
    \end{small}
    \label{tab:compare_huang}

\end{table}
\textbf{\textit{\toolname{} trains more robust LSTMs than CNNs trained by \citet{DBLP:conf/emnlp/HuangSWDYGDK19}.}} 
\Tblref{tab:compare_huang} shows that \toolname{} results in models with higher accuracy ($+1.6$), HF Acc. ($+1.1$), \Verifiedshort{} ($+28.8$), and \Exhaustiveshort{} ($+3.4$) than those produced by \citet{DBLP:conf/emnlp/HuangSWDYGDK19}.

\paragraph{Effectiveness of \toolname{}-A3T} We can apply the idea of A3T to \toolname{}, extending \toolname{} to abstract any subset of the given perturbation space and to augment the remaining perturbation space. We show the effectiveness of this extension in the appendix.

We evaluate \toolname{}-A3T on the same perturbation spaces as we do for A3T. For each perturbation space, \toolname{}-A3T has four instantiations: abstracting the whole perturbation space (downgraded to \toolname{}), abstracting the first perturbation space ($\{(\Tdelword{}, 2)\}$ or $\{(\Tinsword{}, 2)\}$), abstracting the second perturbation space ($\{(\Tsubword{}, 2)\}$), and augmenting the whole perturbation space. We use enumeration for augmenting. We do not test the last instantiation because enumeration the whole perturbation space is infeasible for training. We further evaluate the trained models on different perturbation sizes, i.e., $\{(\Tdelword{}, \delta), (\Tsubword{}, \delta)\}$ and $\{(\Tinsword{},\delta), (\Tsubword{},\delta)\}$ with $\delta=1,2,3$.

\begin{table}[t]
    \centering
    \caption{Results of different instantiations of \toolname{}-A3T on SST2 dataset. }
    \setlength{\tabcolsep}{3pt}

    \begin{small}
    \begin{tabular}{l *2{ccc}}
        \toprule
         & \multicolumn{3}{c}{\scriptsize $\{(\Tdelword{}, 2),(\Tsubword{},2)\}$} & \multicolumn{3}{c}{\scriptsize $\{(\Tinsword{}, 2),(\Tsubword{},2)\}$} \\ 
         
         \cmidrule(lr){2-4} \cmidrule(lr){5-7} 
        Train & {\small Acc.} & {\scriptsize \Verifiedshort{}} & {\scriptsize \Exhaustiveshort{}} & {\small Acc.} & {\scriptsize \Verifiedshort{}} & {\scriptsize \Exhaustiveshort{}}  \\
        \midrule
        Abs-fir & \textbf{83.2}  & 15.1 & 68.6 & 82.6 & 15.9 & \textbf{66.7} \\
        Abs-sec & 81.4  & 71.1 & 75.8 & \textbf{83.0} & 48.8 & 65.4 \\
        \toolname{} & 82.5 & \textbf{72.5} & \textbf{77.0} & 80.2 & \textbf{55.4} & 64.0\\
        \bottomrule
    \end{tabular}
    \end{small}
    \label{tab:arc_a3t}

\end{table}
\textbf{\textit{Different instantiations of \toolname{}-A3T win for different perturbation spaces.}}
\Tblref{tab:arc_a3t} shows the results of different instantiations of \toolname{}-A3T. For $\{(\Tdelword{}, 2), (\Tsubword{}, 2)\}$, abstracting the first perturbation space ($\{(\Tdelword{}, 2)\}$) achieves the best accuracy and abstracting the whole perturbation space (\toolname{}) achieves the best \Verifiedshort{} and \Exhaustiveshort{} For $\{(\Tinsword{},2), (\Tsubword{},2)\}$, abstracting the first perturbation space ($\{(\Tinsword{}, 2)\}$) achieves the best \Exhaustiveshort{}, abstracting the second perturbation space ($\{(\Tsubword{}, 2)\}$) achieves the best accuracy, and abstracting the whole perturbation space (\toolname{}) achieves the best \Verifiedshort{}

\begin{figure}[t]
    \centering
    % \aws{y axis: "size of perturbation space"}
     \includegraphics[width=\columnwidth]{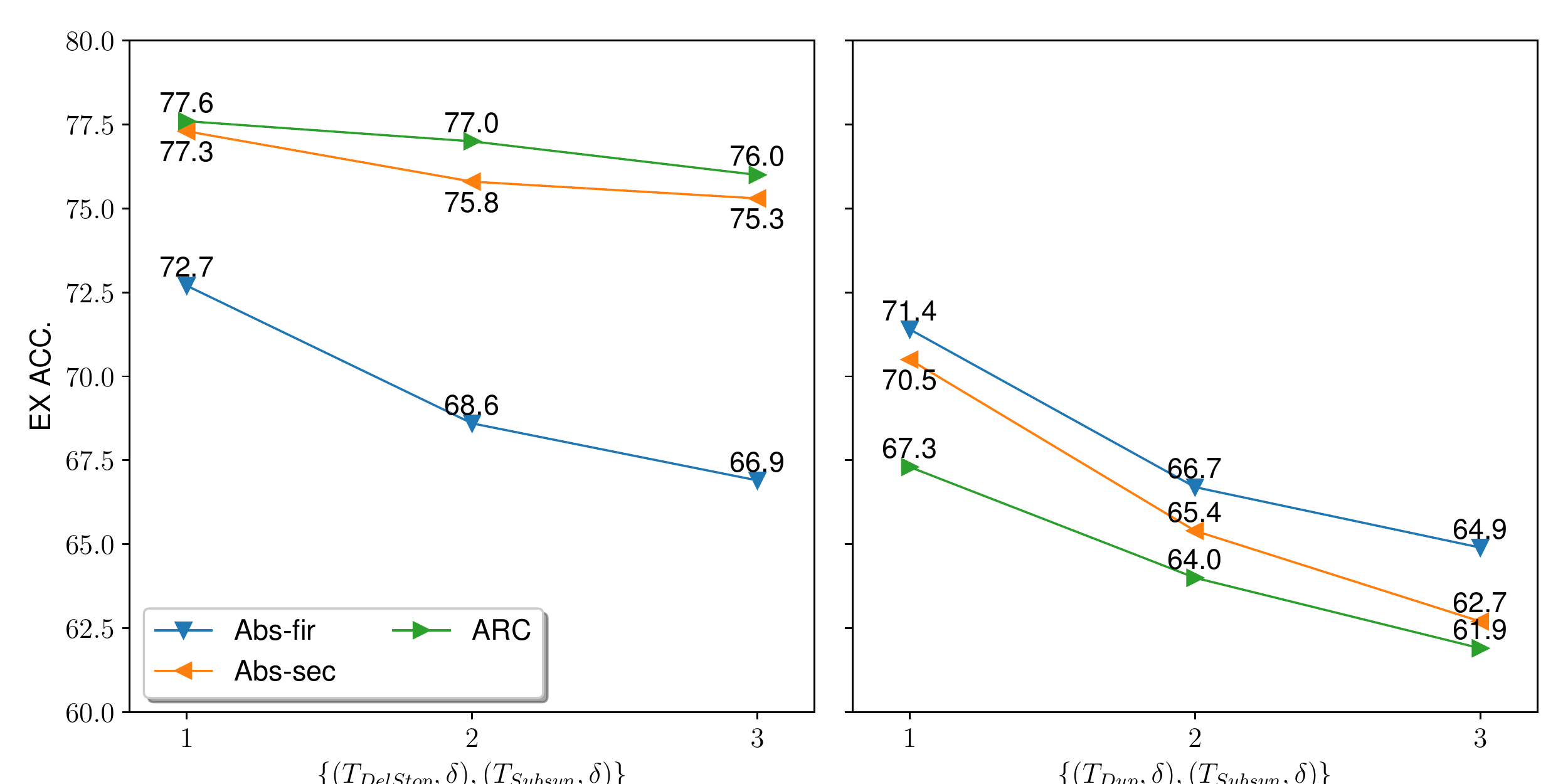}
    \caption{\Exhaustiveshort{} as $\delta$ varies from 1 to 3 for $\{(\Tdelword{},\delta), (\Tsubword{},\delta)\}$ and $\{(\Tinsword{},\delta), (\Tsubword{},\delta)\}$.}
    \label{fig:ARCA3T1}
\end{figure}

\begin{figure}[t]
    \centering
    % \aws{y axis: "size of perturbation space"}
     \includegraphics[width=\columnwidth]{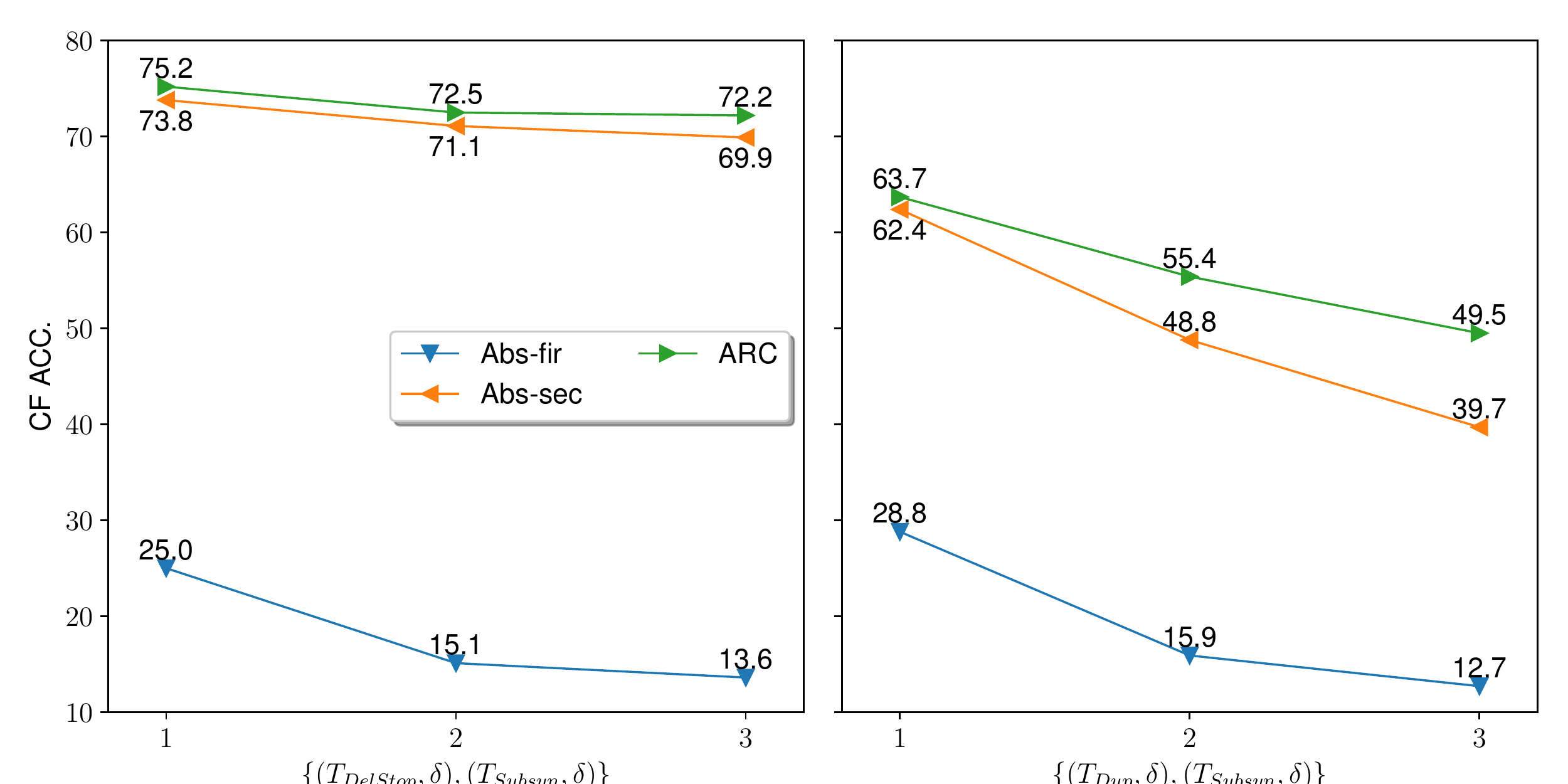}
    \caption{\Verifiedshort{} as $\delta$ varies from 1 to 3 for $\{(\Tdelword{},\delta), (\Tsubword{},\delta)\}$ and $\{(\Tinsword{},\delta), (\Tsubword{},\delta)\}$.}
    \label{fig:ARCA3T2}
\end{figure}

\Twofigref{fig:ARCA3T1}{fig:ARCA3T2} show the \Exhaustiveshort{} and \Verifiedshort{} for $\{(\Tdelword{},\delta), (\Tsubword{},\delta)\}$ and $\{(\Tinsword{},\delta), (\Tsubword{},\delta)\}$, as $\delta$ varies from 1 to 3.
\toolname{} achieves the best \Exhaustiveshort{} and \Verifiedshort{} for $\{(\Tdelword{},\delta), (\Tsubword{},\delta)\}$ and the best \Verifiedshort{} for $\{(\Tinsword{},\delta), (\Tsubword{},\delta)\}$.
Abstracting the first perturbation space ($\{(\Tinsword{},\delta)\}$) achieves the best \Exhaustiveshort{} for $\{(\Tinsword{},\delta), (\Tsubword{},\delta)\}$.
Notice that the abstraction approach proposed by \toolname{} enables the abstracting of $\{(\Tinsword{},\delta)\}$.

\end{document}